\documentclass[journal]{IEEEtran}
\usepackage[usenames]{color}
\usepackage{hyperref}

\let\em\relax
\DeclareTextFontCommand{\em}{\it}
\let\emph\relax
\DeclareTextFontCommand{\emph}{\it}

\usepackage{graphicx}
\usepackage{bm} 
\usepackage{epsfig} 
\usepackage{mathptmx} 
\usepackage{times} 
\usepackage{amsmath} 
\usepackage{amssymb}  
\usepackage{algorithm}
\usepackage{algorithmic}
\usepackage{comment}
\usepackage{cite}
\usepackage{enumerate}

\def\Vec#1{\mbox{\boldmath $#1$}}
\usepackage[hang,small,bf]{caption}
\usepackage[subrefformat=parens]{subcaption}
\title{Unsupervised Multimodal Word Discovery \\based on Double  Articulation Analysis \\with Co-occurrence Cues}

\author{%
    Akira Taniguchi, 
    Hiroaki Murakami, 
    Ryo Ozaki, and 
    Tadahiro Taniguchi,~\IEEEmembership{Member,~IEEE}
    \thanks{A. Taniguchi and T. Taniguchi are with the College of Information Science and Engineering, Ritsumeikan University, 1-1-1 Noji Higashi, Kusatsu, Shiga 525-8577, Japan {\tt\small a.taniguchi@em.ci.ritsumei.ac.jp, taniguchi@em.ci.ritsumei.ac.jp}}
    \thanks{H. Murakami, R. Ozaki are with the Graduate School of Information Science and Engineering, Ritsumeikan University, 1-1-1 Noji Higashi, Kusatsu, Shiga 525-8577, Japan {\tt\small murakami.hiroaki@em.ci.ritsumei.ac.jp, ryo.ozaki@em.ci.ritsumei.ac.jp}}
}

\begin{document}

\markboth{IEEE Transactions on Cognitive Developmental Systems,~Vol.~00, No.~0, XX~2023}%
{Shell \MakeLowercase{\textit{et al.}}: Bare Demo of IEEEtran.cls for IEEE Journals}

\maketitle

\begin{abstract}
Human infants acquire their verbal lexicon with minimal prior knowledge of language based on the statistical properties of phonological distributions and the co-occurrence of other sensory stimuli.
This study proposes a novel fully unsupervised learning method for discovering speech units using phonological information as a distributional cue and object information as a co-occurrence cue. 
The proposed method can acquire words and phonemes from speech signals using unsupervised learning and utilize object information based on multiple modalities—vision, tactile, and auditory—simultaneously. 
The proposed method is based on the nonparametric Bayesian double articulation analyzer (NPB-DAA) discovering phonemes and words from phonological features, and multimodal latent Dirichlet allocation (MLDA) categorizing multimodal information obtained from objects. 
In an experiment, the proposed method showed higher word discovery performance than baseline methods.
Words that expressed the characteristics of objects (i.e., words corresponding to nouns and adjectives) were segmented accurately.
Furthermore, we examined how learning performance is affected by differences in the importance of linguistic information.
Increasing the weight of the word modality further improved performance relative to that of the fixed condition.
\end{abstract}


\IEEEpeerreviewmaketitle

\begingroup
\renewcommand{\thefootnote}{}
\endgroup

\section{Introduction} \label{sec:Introduction}

\IEEEPARstart{H}UMAN infants can acquire their verbal lexicon with minimal prior knowledge based on the statistical properties of phonological distributions and co-occurrence of other sensory stimuli~\cite{saffran1996statistical,pelucchi2009statistical,saffran2020statistical}.
Regarding the importance of fundamental statistical regularity in the lexical acquisition by infants, Saffran et al. observed that there are three key elements; (1)~distributional cues, (2)~co-occurrence cues, and (3)~prosodic cues~\cite{saffran1996word}.
Here, distributional cues are the statistical relationships regarding the phonological information in utterances, and co-occurrence cues are the information provided by the sensory stimulus that co-occurs with a specific utterance. 
Prosodic cues are information (such as intonation) included in utterances and the silent sections generated between utterances.
A study of infant statistical learning~\cite{saffran2018infant} reported that infants are sensitive to statistical regularities in many domains such as speech, music, behavior, and spatial vision.
Statistical learning mechanisms allow infants to discover statistical regularities in the environment, such as the words contained in utterances.

As co-occurrence cues are described as one of the important factors in lexical acquisition, infants observe various other types of sensory stimulus simultaneously when hearing speech~\cite{younger1985segregation}.
Humans can classify things into categories by observing various types of sensory information from early in childhood and these categories play an important role in human cognitive function~\cite{ashby2005human}.
Additionally, it is considered that infants can change the type of important information to which they are attending depending on the progress of learning~\cite{jusczyk1999beginnings,kuhl2006infants}.
However, how to specifically change the importance of given information remains an {open} issue. 
Therefore, this study focuses on the importance of co-occurrence cues in the lexical acquisition process and the effect of changes in their importance for efficient learning.

There is an approach that aims to elucidate the lexical acquisition process by imitating the function of humans and expressing it via machine-learning methods~\cite{Roy2002,goldwater2009bayesian,rasanen2015unsupervised,lee2015unsupervised, taniguchi2016nonparametric,chen2019audio}.
This type of approach is referred to as a constructive approach.
The findings obtained from these computational models that partially imitate human language learning functions contain clues for elucidating human language learning functions. 
This approach can be used to develop robots with functions that more closely approximate those of humans.
In this study, we focused on the language learning function of infants, who can discover voice units from spoken utterances.
This function is expressed as a speech unit discovery method via unsupervised learning that does not use labeled data for its machine learning~\cite{taniguchi2016nonparametric}.
In speech unit discovery using computational models, words and phonemes are often considered as speech units~\cite{lee2015unsupervised, taniguchi2016nonparametric, chen2019audio}.
There have been various approaches in this area of research such as those that assume that both words and phonemes are speech units~\cite{lee2015unsupervised, taniguchi2016nonparametric}, or those that focus on only words~\cite{kamper2016unsupervised} or only phonemes~\cite{ondel2016variational}.
However, the segmentation accuracy of the latter methods is low due to several factors.
For example, an over-segmentation of words can occur based on the recognition error of phonemes. 
Therefore, it is important to consider both words and phonemes as speech units, using a \textit{double articulation analyzer (DAA)}.

Several computational models for the discovery of speech units have been proposed to utilize other types of information that co-occur with linguistic information~\cite{nakamura2014mutual, ataniguchi2017spcoslam, taniguchi2018unsupervised, Chrupa_a_2017, harwath2017learning, wang2020dnn}.
There are various types of co-occurrence cues and the relationships between co-occurrence cues and linguistic information {that} have some explanatory value.
Many studies have assumed a set of images and linguistic captions that explain the image~\cite{Chrupa_a_2017, harwath2017learning, wang2020dnn}.
With such methods, the accuracy of speech unit discovery is improved by learning the association between the object in the image and the speech unit, as compared with cases in which no image is given.
However, these studies are not aimed at lexical acquisition and use only one type of co-occurrence information.
When considering the imitation of human statistical learning, it is desirable to use co-occurrence information based on multiple types of sensory stimuli simultaneously.
There has been some previous research that meets this requirement~\cite{nakamura2014mutual}.
In this study, multiple modalities, specifically, the image of the object, the tactile feel when the object is grasped, and the sound when the object is shaken, are handled as co-occurrence cues for the spoken utterances that express the characteristics of the object. 
Similar studies~\cite{taniguchi2018unsupervised, ataniguchi2017spcoslam} used the position of a robot and the image at its place as co-occurrence cues for spoken utterances that express places. 
However, these studies assume that phonemes and syllables have already been acquired, and thus cannot conclude that lexical acquisition is completely achieved via unsupervised learning.

\begin{figure}[tb]
  \begin{center}
    \includegraphics[width=\linewidth]{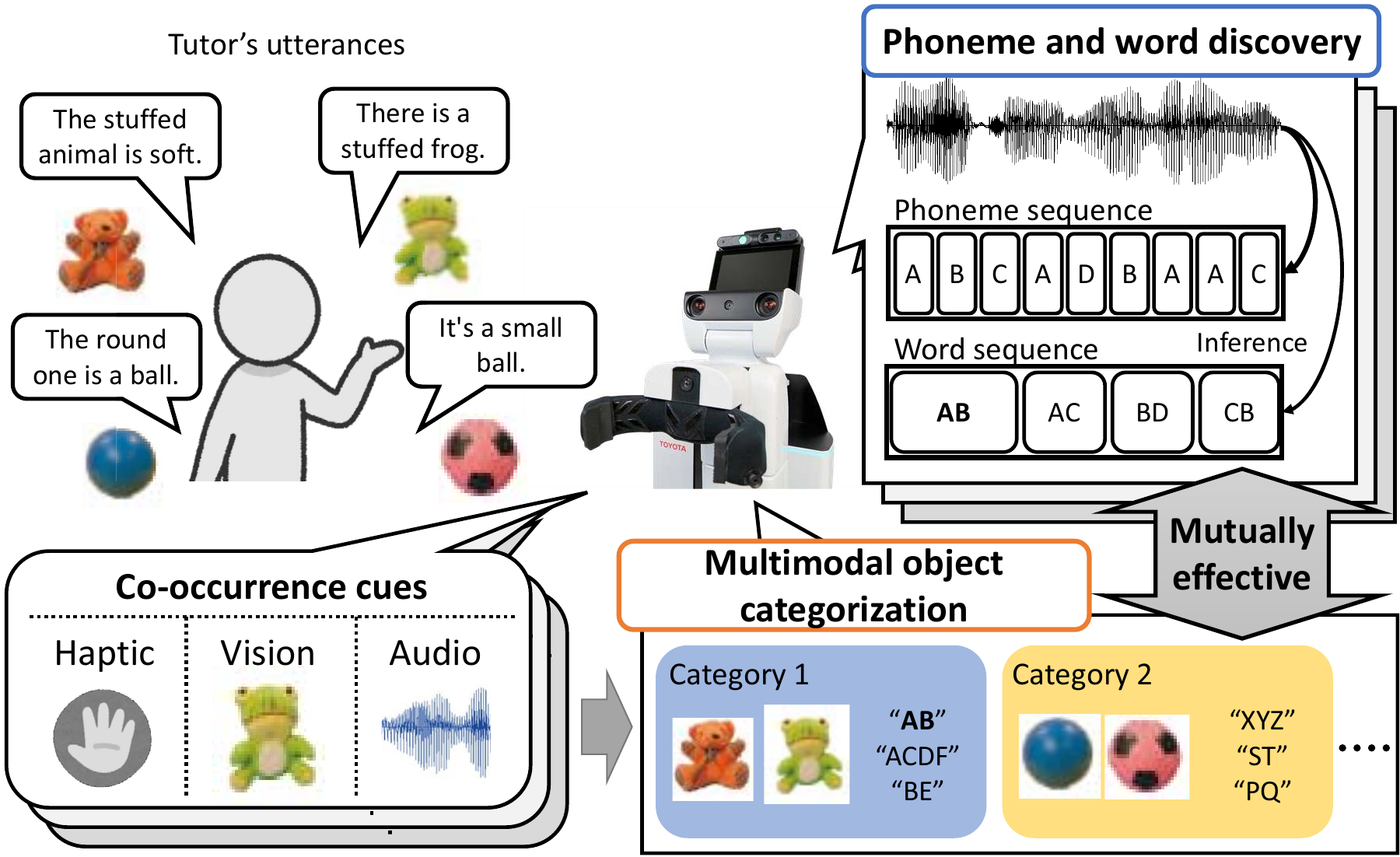}
    \caption{Overview of this study.
    Phonemes and words are discovered by simultaneously using human utterances representing objects' characteristics and the multimodal object information that co-occurs with utterances.
    }
    \label{fig:abst}
  \end{center}
\end{figure}

In this study, we propose \textbf{co-occurrence DAA}, a novel fully unsupervised learning method that discovers phonemes and words using phonological information as distributional cues and multiple other forms of sensory information as co-occurrence cues.
The proposed method is based on the probabilistic generative model, \textbf{HDP-HLM+MLDA}, which integrates a hierarchical Dirichlet process hidden language model (HDP-HLM)~\cite{taniguchi2016nonparametric} and multimodal latent Dirichlet allocation (MLDA)~\cite{nakamura2011grounding}.
{The integration of the two models is based on the concept of Symbol Emergence in Robotics Tool Kit (SERKET)~{\cite{nakamura2018serket,taniguchi2020neuroSERKET}} using the sampling importance resampling (SIR) method~\cite{rubin1988using}.}
SERKET is the theoretical framework for the integration of probabilistic generative models, and the construction and demonstration of the integrated inference algorithm with SIR is one of the novelties of this study.
The overview of this study is presented in Fig.~\ref{fig:abst}.
We investigate how co-occurrence cues affect phoneme and word discovery performance and compare learning results depending on the importance of co-occurrence cues.
Hence, the main contributions are as follows:
\begin{enumerate}
    \item We construct a fully unsupervised learning method that uses not only distributional cues but also co-occurrence cues for phoneme and word discovery.
    \item We show using co-occurrence cues improves word segmentation performance.
    {(mainly shown in Table~\ref{tab:ex_1_last_ARI} of Experiment  1)}
    \item We suggest that co-occurrence cues regarding objects facilitate the discovery of words regarding objects. 
    {(Mainly shown in Figure~\ref{fig:segment1} of Experiment 1)} 
    \item Performances of word discovery and object categorization are further improved by increasing the weight of the word modality.
    {(Mainly shown in Table~\ref{tab:ex_2_last_ARI} and Figure~\ref{fig:segment2} of Experiment 2)} 
\end{enumerate}
This study is novel, and its results can be applied in basic research domains focused on a better understanding of language acquisition, as well as in a variety of practical applications using language (e.g., human-robot interactions).
Here, we open the source code  of the proposed method and speech dataset on GitHub\footnote{\url{https://github.com/a-taniguchi/NPB-DAA-MLDA}}.

The remainder of this paper is structured as follows.
First, Section~{\ref{sec:related_work}} describes previous research that examines lexical acquisition and categorization by infants, computational models for phoneme and word discovery, and word discovery methods using co-occurrence cues. 
Next, Section~{\ref{sec:preparation}} introduces the conventional methods, MLDA and NPB-DAA, as the background for the proposed method.
Section~{\ref{sec:proposed_method}} describes the proposed method.
Then, we describe the experiments in Sections~\ref{sec:ex_1} and \ref{sec:ex_2}.
Finally, we provide a conclusion and directions for future work in Section~{\ref{sec:conclusion}}.

\section{Related work} \label{sec:related_work}

We describe four types of related work considered in this study.
Section~\ref{sec:related_work:human} describes studies on lexical acquisition and categorization in infants.
Section~\ref{sec:related_work:constructive} describes how the constructive approach has been applied to lexical acquisition.
Section~\ref{sec:related_work:computation} describes unsupervised speech unit discovery methods that work from speech data only.
Section~\ref{sec:related_work:co-occurrence} describes word discovery methods that use co-occurrence cues.

\subsection{Lexical Acquisition and Categorization in Infants} \label{sec:related_work:human}

Various approaches have been studied to date to elucidate the factors influencing lexical acquisition in infants~\cite{choi2020preverbal,saffran2018infant,saffran2020statistical,saffran1996word,jusczyk1999beginnings,kuhl2006infants}. 
In general, it is believed that infants discover words using speech statistical distribution information.
For example, in one study on the role of distributional cues in word segmentation~\cite{saffran1996word}, a word segmentation experiment was conducted using an artificial language and adult subjects.
These experiments suggested that distributional cues play an important role in the early word segmentation of language learners.
{However, word segmentation using distributional cues alone is difficult owing to biases and deficiencies in observed words during language learning. 
Moreover, the language input may vary due to factors such as dialect, accent, speaking rate, and external environment and context changes.}
Therefore, Saffran et al. contended that not only distributional cues but also multimodal sensory information such as prosodic and co-occurrence cues, are important in lexical acquisition~\cite{saffran1996word}.

Although experiments with infants have reported some results, there are some {remaining} issues.
Pelucchi et al.~\cite{pelucchi2009statistical} showed that infants are sensitive to syllable transition probabilities in natural language stimuli and that statistical learning is robust enough to support lexical acquisition in the real world. 
One study on word segmentation for infants learning English~\cite{jusczyk1999beginnings} focused on accents during speech and showed that distributional cues play an important role in the early stages of word segmentation learning. 
Therefore, it is considered that infants change the importance assigned to each source of information depending on their progress in the language learning process.
Kuhl et al.~\cite{kuhl2006infants} investigated whether the importance of information affects perceptual accuracy as learning progresses.
Previous experimental evaluations~\cite{pelucchi2009statistical,jusczyk1999beginnings,kuhl2006infants} are widely used to assess language learning.
However, because {these} behavioral experiments were conducted after learning, they are susceptible to various external factors. 
For example, they cannot observe the dynamic progress of learning and similarly cannot target adult subjects.
Choi et al.~\cite{choi2020preverbal} proposed using the measurement results of electroencephalograms worn during the experiment to overcome these problems.
However, this introduces new issues, such as the costly nature of electroencephalogram measurements. 
Therefore, a constructive approach based on a computational model, as introduced in Section~\ref{sec:related_work:constructive}, is useful.

Co-occurrence cues have been described as one of the most important factors in lexical acquisition.
Infants can observe various other types of sensory stimuli and hear speech simultaneously.
In fact, it has been reported that 10-month-old infants can discover simple categories from visual information~\cite{younger1985segregation}.
In this way, it has been observed that humans can classify things into categories by observing various types of sensory information from early childhood and that these categories play an important role in human cognitive function~\cite{ashby2005human}.
One study of language acquisition by infants reported that infants tend to understand a word as the name of a category to which the target object belongs, rather than as a proper noun~\cite{markman1984children}.
However, the details of nature and the relationship between lexical acquisition and category formation in infant development remain unresolved and controversial~\cite{Yin2015}.
Additionally, Okanoya et al. presented a hypothesis of string-context mutual segmentation in language evolution~\cite{Okanoya2007}.
{Our proposed method can be interpreted as a computational model that represents the functions an agent should have to embody this hypothesis.}

\subsection{Constructive Approach to Lexical Acquisition} \label{sec:related_work:constructive}

Several studies have used a constructive approach to imitate the functions and developmental processes of humans and express them as machine learning methods to further elucidate the process. 
The advantage of this approach is that it can be analyzed relatively easily, with the learning results and parameters as a computational process.
Of course, it cannot be concluded at this stage whether the machine learning methods used in this approach accurately represent the human developmental process.
However, the knowledge obtained from methods that have functions similar to human lexical acquisition can and will be used to understand the human lexical acquisition process better.
In addition, the knowledge obtained from machine learning methods will be used to develop robots that have functions closer to humans.

In recent years, several studies on lexical acquisition using a constructive approach have been conducted~\cite{lee2015unsupervised, rasanen2015unsupervised, taniguchi2016nonparametric, chen2019audio, kawakami-etal-2019-learning}.
Several machine learning methods imitate the lexical acquisition process of infants using a constructive approach, such as the phoneme and word discovery method known as the nonparametric Bayesian double articulation analyzer~(NPB-DAA).
NPB-DAA is an unsupervised-learning method based on Bayesian inference, which assumes that the time-series data has a double-articulation structure.
Here, double articulation refers to a structure in which the time-series data can be segmented into a certain unit, and each unit can also be segmented into chunks.
For example, human utterances can be segmented into units of words, and each word can be segmented into another unit called phonemes, thus it has a double articulation structure.
One of the main features of NPB-DAA is that not only can words and phonemes be acquired through fully unsupervised learning, but it can also be applied to relatively small datasets.
Therefore, in this study, NPB-DAA was used as the base of the proposed method, and the outline is described in Section~\ref{sec:preparation}.

In the constructive approach for lexical acquisition, many studies only used speech signals.
For example, a lexical discovery method~\cite{lee2015unsupervised} that supports variously changing phoneme and word expressions extended adapter grammars~\cite{johnson2007adaptor}, which is the nonparametric Bayes morphological analysis.
However, owing to limitations in the noisy-channel model used for modeling variability, sufficient lexical acquisition performance could not be achieved in this study.
Studies on syllables are common in the field of lexical acquisition and speech recognition tasks by infants~\cite{jusczyk1999beginnings, cutler1988role, eimas1999segmental}.
Alternatively, machine-learning studies focusing on syllables are relatively rare because it is difficult to achieve unanimity in the detection and definition of syllables~\cite{rasanen2015unsupervised}.
Previous experimental results, however, have shown high accuracy, especially in word segmentation, which is useful for future reference in the approach of word discovery.
Audio Word2vec~\cite{chen2019audio}, which is an extension of Word2Vec~\cite{mikolov2013distributed} applied to speech data, segments speech utterances at the word level and then converts those words into vector representations.
However, in these studies, speech unit discovery was performed using only the information obtained from the speech signal, and other sensory co-occurrence cues  were not used.

\subsection{Unsupervised Speech Unit Discovery Methods from Speech Data Only} \label{sec:related_work:computation}

Lexical acquisition by unsupervised learning is cost-effective because it does not require a large amount of labeled data to be prepared for learning.
Studies on discovering units, such as words, from spoken utterances using unsupervised learning have been conducted using various approaches~\cite{lee2012nonparametric,kamper2016unsupervised,ondel2016variational,glarner2018full,chorowski2019unsupervised,van2020vector}.
The main purpose of these previous studies was to enable automatic speech recognition to be applied to languages with few resources for learning, rather than imitating the infant statistical learning process via the constructive approach, as was introduced in the previous section. 
Kamper et al.~\cite{kamper2016unsupervised} proposed a method that used acoustic word embeddings in word segmentation by unsupervised learning; however, they did not explicitly deal with phoneme or syllable segmentation but focused only on word segmentation.

In research on speech unit discovery using unsupervised learning, a method based on a variational auto-encoder (VAE) was proposed~\cite{glarner2018full,chorowski2019unsupervised,van2020vector}.
The Bayesian hidden Markov model VAE~\cite{glarner2018full} is a speech unit discovery method that extends VAE by embedding a Bayesian framework in the hidden Markov model VAE~\cite{ebbers2017hidden}.
Specifically, by assuming the Dirichlet process (DP) as a prior distribution for the distribution of speech units, it is possible to automatically infer the total number of speech units.
Our proposed method can also automatically infer the number of phonemes and words using DP.
Neural network-based speech representation learning~\cite{chorowski2019unsupervised} can obtain discrete representations using vector quantized VAE~(VQ-VAE)~\cite{oord2017neural}.
Additionally, it can retain a significant amount {of} linguistic information and the invariance of the speaker.
Niekerk et al.~\cite{van2020vector} investigated the usefulness of vector quantization in learning representations that separate speech content and the characteristics peculiar to the speaker. 
Recently, wav2vec-U~\cite{Baevski2021}, a method for unsupervised speech recognition using phonemized unlabeled text via generative adversarial networks~\cite{Goodfellow2014}, has been developed.
These prior studies showed high performance in word discovery.
However, the models used did not use purely unsupervised learning from only speech data and functioned with preliminary assumptions such as the use of texts or codebooks of phonemes.
Therefore, they are different from developmental models that imitate the lexical acquisition processes of infants with the aim of understanding their functions.

\subsection{Word Discovery Methods Using Co-occurrence Cues} \label{sec:related_work:co-occurrence}

Some studies have taken the approach of using co-occurrence cues other than utterances simultaneously in word discovery~\cite{nakamura2014mutual,Chrupa_a_2017,harwath2017learning,taniguchi2018unsupervised,wang2020dnn}.
The motivations for using information other than utterances include improving the performance of word discovery and providing linguistic connections to co-occurrence cues.
Nakamura et al.~\cite{nakamura2014mutual} proposed a word discovery method using multimodal sensor data that can be observed from an object as co-occurrence cues in word discovery.
In this study, object categorization was performed using utterance information and multimodal sensor data, and each word was linked to the object category.
Specifically, the nested Pitman-Yor language model (NPYLM)~\cite{mochihashi2009bayesian} was used for word discovery, and multimodal latent Dirichlet allocation (MLDA)~\cite{nakamura2011grounding} was used for object categorization.
NPYLM can discover words via unsupervised learning, but because the input needs to be in text format, it is assumed that phonemes or syllables can be recognized.
MLDA is an unsupervised categorization method that can handle multiple modalities simultaneously (for details, see Section~\ref{sec:preparation:mlda}).
In addition, SpCoA++~\cite{taniguchi2018unsupervised}, SpCoSLAM~\cite{ataniguchi2017spcoslam}, and ReSCAM~\cite{Sagara2021} can learn the place category and the lexicon based on the syllable recognition lattices and the sensor observations about the place as co-occurrence cues.
These studies reported that simultaneous learning of categories and the lexicon improves accuracy in both word segmentation and categorization.
In addition to the situational context co-occurring with speech, leveraging a top-down grammar learning process also improves word segmentation performance~\cite{Gaspers2017}.
However, these studies assumed a certain level of prior knowledge of phonemes and syllables.
In our study, we proposed a method that can simultaneously detect phonemes and words by referring to the above approach.
This can be interpreted as a machine learning method that imitates the process of an infant acquiring lexicon.

Although not focused on lexical acquisition or speech unit discovery, some studies link speech and images~\cite{Chrupa_a_2017, harwath2017learning, wang2020dnn}.
The visually linked speech recognition model projects speech utterances and images into a common semantic space~\cite{Chrupa_a_2017}.
The method for finding a word and associating that word with an object in an image uses both the image and its speech caption~\cite{harwath2017learning}.
Such a method does not use existing speech recognition devices or prior linguistic annotations.
However, word segmentation using this method is insufficient for sections of speech that are not sufficiently associated with images.
The hybrid model comprising a deep neural network and a hidden Markov model discovers words from images representing objects and their audio captions~\cite{wang2020dnn}.
Because the above model does not consider word-level information, there remains the problem of confusing different words that share phonemes.

\section{{Foundational Methods}} \label{sec:preparation}

The proposed method is based on NPB-DAA, which is an unsupervised phoneme and word discovery method from phonological features, and MLDA, which is an unsupervised object categorization method for multimodal information obtained from objects.
This section provides an overview of two foundational methods as MLDA in Section~\ref{sec:preparation:mlda} and NPB-DAA in Section~\ref{sec:preparation:npbdaa}.
{The integrated model and inference in the proposed method are described in Section~\ref{sec:proposed_method}.}

MLDA can discover the category of an object by clustering observation data obtained when a robot sees, grasps, and makes sounds with various objects, without requiring hand-labeled category labels.
NPB-DAA can segment speech into phonemes and words based on only features of the speech, without any hand-labeled segmentation boundary.
Therefore, MLDA and NPB-DAA may not always produce accurate results, and different variations of results may be obtained depending on the ambiguity of the observation data.
For example, a robot may observe various objects including coins and buttons, but MLDA may classify them into the same category owing to their visual similarity.
On the other hand, humans may verbally say ``This is a coin found on the street corner'' or ``A round button fell'' and NPB-DAA may segment the speech data into the words such as ``/coi/ /nf/'' and ``/but/ /to/ /nf/''.
In our study, we integrate these methods by connecting the features of objects and speech in robots and making them refer to each other's learning results, to teach the robot that ``coin'' and ``button'' are different words and belong to different categories.

\subsection{Multimodal Latent Dirichlet Allocation: MLDA} \label{sec:preparation:mlda}

\begin{figure}[!t]
     \centering
     \begin{subfigure}[b]{\linewidth}
         \centering
         \includegraphics[width=\textwidth]{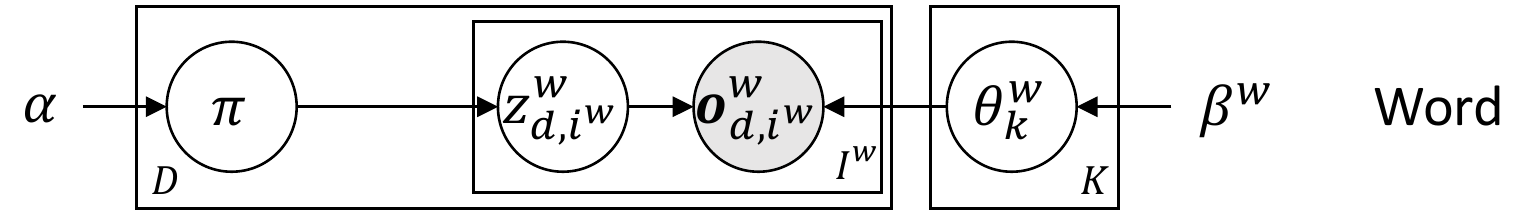}
         \caption{LDA}
         \label{fig:graphical_model_lda}
     \end{subfigure}
     \\
     \hfill
     \\
     \begin{subfigure}[b]{\linewidth}
         \centering
         \includegraphics[width=\textwidth]{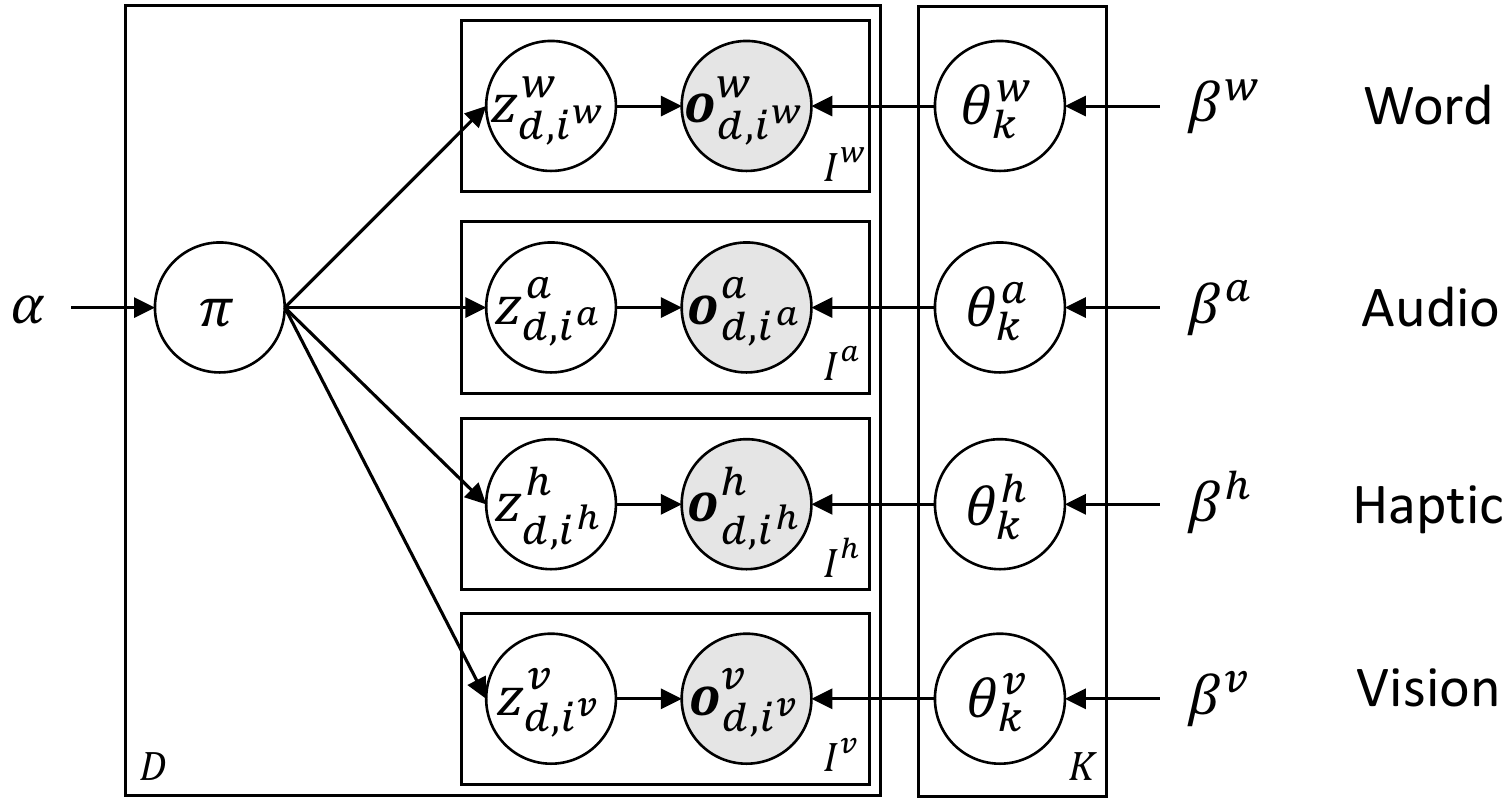}
         \caption{MLDA}
         \label{fig:graphical_model_mlda}
     \end{subfigure}
        \caption{
            The graphical models of {(a) LDA and (b)} MLDA (a version consisting of four modalities; From the top: word, audio, haptic, and vision modalities).
        }
        \label{fig:mlda}
\end{figure}

Nakamura et al.~\cite{nakamura2011grounding} extended the latent Dirichlet allocation (LDA)~\cite{blei2003latent} as an MLDA that can generalize multiple types of sensory observation simultaneously to enable object categorization using multimodal data.
For instance, images may have certain colors and shapes while audio data may have specific acoustic features. 
By extracting common categories across these different modalities, MLDA can understand the relationships between multiple modalities. 
The word distributions are obtained based on the observed frequencies of words for each category. 
It has been shown that object categorization using MLDA is closer to human senses than categorization using a single modality.
For more details on this approach, refer to the original {MLDA paper}~\cite{nakamura2011grounding}.


LDA~\cite{blei2003latent} is a prominent representative method for topic modeling. 
The original LDA was developed to estimate the potential topic, that is, the latent category, for each word from text documents including many sentences.
Figure~\ref{fig:graphical_model_lda} shows the graphical model of LDA.
$ o^{w}_{d,i^{w}} $ is a bag-of-words representation in a document.
{The bag-of-words representation is a way to represent text as a collection of words and their frequencies, without considering the order in which they appear\footnote{This is also the case in MLDA, where descriptions, that is, sequences of words, about an object given by the instructor is used as a word histogram for each object.}}.
$ \theta^{w} $ represents the word appearance probability for each category, and $ \beta^w $ is a hyperparameter of the Dirichlet prior distribution.
$ z^{w} $ refers to the index of the category assigned to each word.
$ \pi $ is a parameter of the multinomial distribution representing the probability of the appearance of each category, and the Dirichlet prior distribution with $ \alpha $ as a hyperparameter is used as the prior distribution of this multinomial distribution.
$D$ is the number of documents, $I^{w}$ is the number of words in a document, and $K$ is the number of topics.

Figure~\ref{fig:graphical_model_mlda} shows the graphical model of MLDA.
Here, the superscripts $ w $, $ a $, $ h $, and $ v $ represent different modalities, indicating linguistic, auditory, tactile, and image information, respectively.
$ o^* $ refers to the features of each modality.
$ \theta^* $ represents the appearance probability of different features for each category in each modality, and each follows the Dirichlet prior distribution with $ \beta^* $ as a hyperparameter.
$ z^* $ refers to the index of the category assigned to each feature in each modality.
$ \pi $ is a parameter of the multinomial distribution, and $ \alpha $ is a hyperparameter of the Dirichlet prior distribution.
The number of objects is $D$, the number of observed features is $I^*$, and the number of categories is $K$.
The features of each modality are represented as bag-of-features.

The collapsed Gibbs sampler is used for MLDA parameter estimation, as shown in Algorithm~\ref{alg:mlda}.
The collapsed Gibbs sampler uses marginalized conditional probabilities on $ \pi $ and $ \theta^m $ regarding the category $ z^m_{di} $ assigned to the $ i $-th feature of the $ d $-th object in the modality $ m $ as follows:
\begin{eqnarray}
    && P(z^{m}_{di}=k \mid \Vec{z}^{-mdi},\Vec{o}^m,\alpha,\beta^m) \nonumber \\
    && \propto (N^{-mdi}_{kd}+\alpha) \frac{N^{-mdi}_{mko^{m}}+\beta^m}{N^{-mdi}_{mk}+{\rm Dim}(m)\beta^m},
    \label{eq:mlda_gibbs}
\end{eqnarray}
where ${\rm Dim}(m)$ is the dimension number of the histogram of modality $m$.
The subtraction subscript in Eq.~(\ref{eq:mlda_gibbs}) indicates that the data in that index is excluded from the histogram.

In addition, $N_{mkdo^{m}}$ is the count number of data assigned to the category $k$ and data $o^m$ of modality $m$ in the $d$-th object.
The count numbers are shown as follows:
\begin{eqnarray}
    N_{mko^{m}} = \sum_{d} N_{mkdo^{m}}, \\
    N_{kd}      = \sum_{m,o^m} N_{mkdo^{m}}, \\
    N_{mk}      = \sum_{d,o^m} N_{mkdo^{m}},
\end{eqnarray}
where $N_{mk}$ is the count number assigned to the category $k$ for each feature in all objects for modality $m$.

\begin{algorithm}[tb]
    \caption{Collapsed Gibbs sampler for MLDA.}
    \label{alg:mlda}                     
    \begin{algorithmic}
        \REPEAT
        \FORALL{$m, d, i$}
        \item $u\leftarrow {\rm random~value~between}~ [0, 1]$
        \FOR{$k\leftarrow 1$ to $K$}
        \item $P[k]\leftarrow P[k-1]+(N^{-mdi}_{kd}+\alpha) \frac{N^{-mdi}_{mko^{m}}+\beta^m}{N^{-mdi}_{mk}+{\rm Dim}(m)\beta^m}$
        \ENDFOR
        \FOR{$k\leftarrow 1$ to $K$}
        \IF{$u<P[k]/P[K]$}
        \item $z^{m}_{di}=k$, break\\
        \ENDIF
        \ENDFOR
        \ENDFOR
        \UNTIL{a predetermined exit condition is satisfied.}
    \end{algorithmic}
\end{algorithm}

Therefore, the global parameters of MLDA can be acquired as the estimation result in the following:
\begin{eqnarray}
    \theta^{m}_{k,o^{m}}&=&\frac{N_{mko^{m}}+\beta^{m}}{N_{mk}+\text{Dim}(m) \beta^m}, \\
    \pi_{d,k}&=&\frac{N_{kd}+\alpha}{\sum_{k} N_{kd} + K\alpha}.
\end{eqnarray}

\subsection{Nonparametric Bayesian Double Articulation Analyzer: NPB-DAA} \label{sec:preparation:npbdaa}

NPB-DAA is an unsupervised phoneme and word discovery method proposed to computationally imitate the lexical acquisition process of human infants.
NPB-DAA uses an acoustic model and a word model to express phonemes and words.
The acoustic model stochastically represents the duration of each phoneme and the acoustic features that make up the phoneme.
The word model consists of a phoneme bigram model and a word dictionary.
The bigram language model is the probability of transitioning to the word that appears after each word, and the word dictionary is the probability of the phonemes that make up each word.
For more details, refer to previous research~\cite{taniguchi2016nonparametric}.

\begin{figure}[tb]
  \begin{center}
    \includegraphics[width=\linewidth]{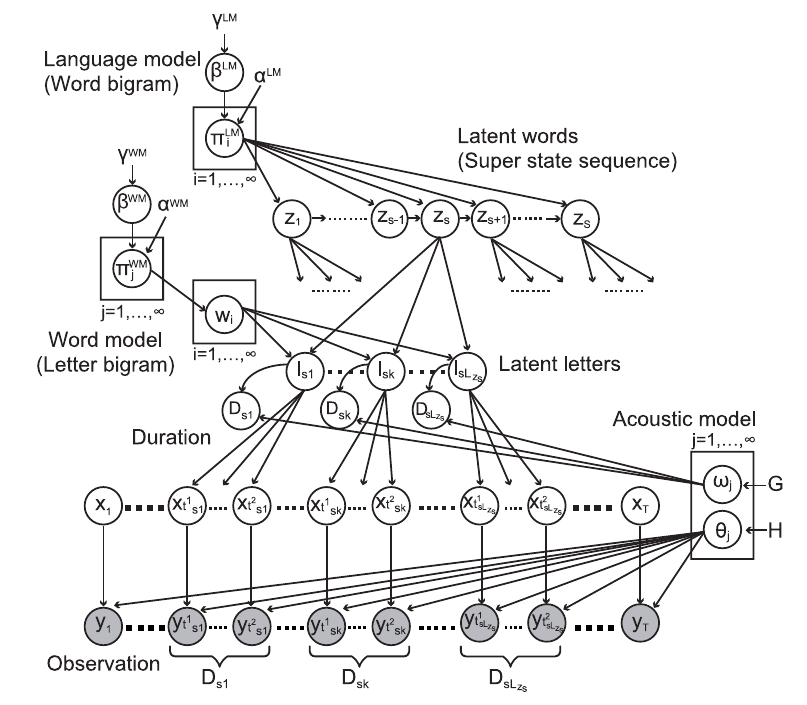}
    \caption{Graphical model representation of HDP-HLM for NPB-DAA~\cite{taniguchi2016nonparametric}.}
    \label{fig:hdphlm}
  \end{center}
\end{figure} 

\begin{figure*}[!tb]
  \begin{center}
    \includegraphics[width=\linewidth]{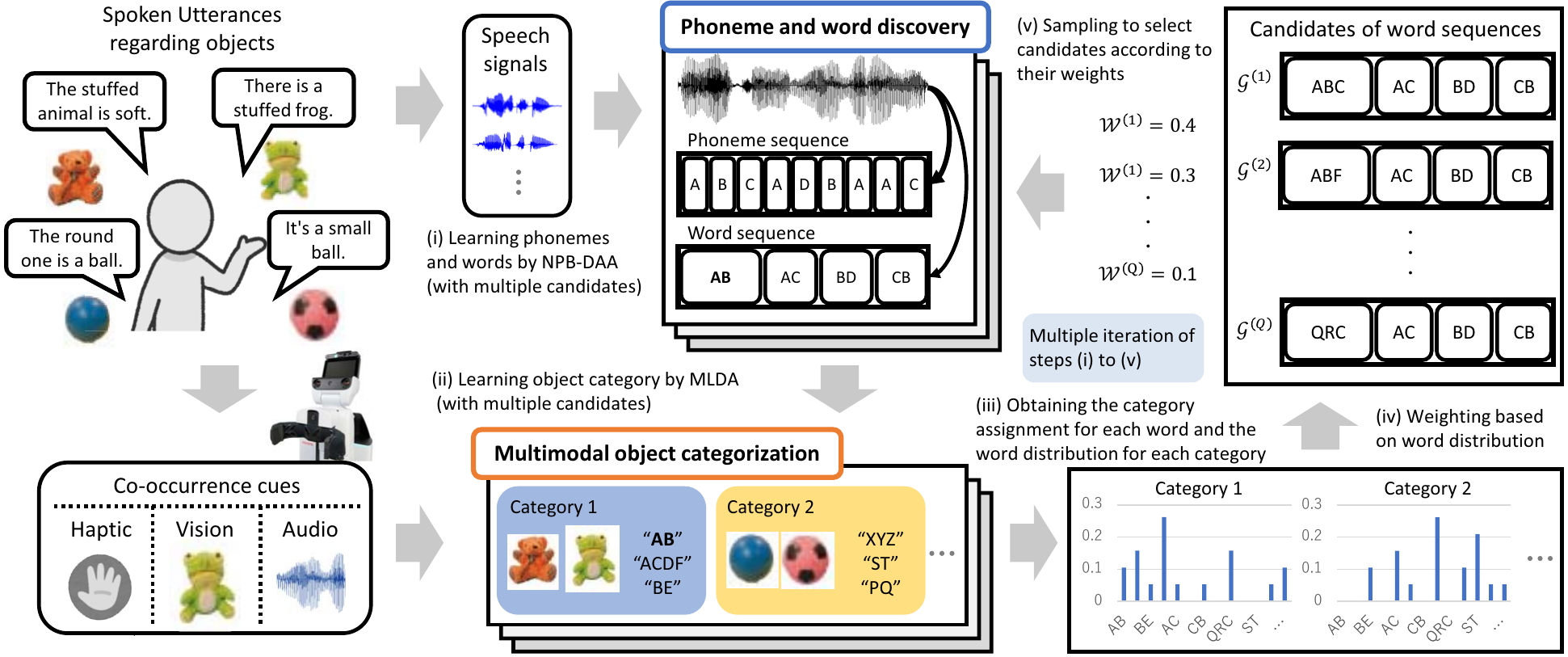}
    \caption{Flow of the iterative estimation of co-occurrence DAA.
    The following procedure was implemented to utilize the co-occurrence of object information and phonological information.
(i) Phonemes and words are learned in each iteration of NPB-DAA; word sequences are estimated for each of the multiple model candidates.
(ii) Object categorization is performed using each candidate of estimated word sequences and multimodal object information.
(iii) The probability distribution of words that are likely to appear in each category is obtained from the categorization results.
(iv) Each model candidate is weighted based on the appearance probability of words in the category assigned to each word included in the word sequences estimated from the utterance.
If the candidate has a higher weight, the words frequently appearing in the category to which the object belongs can be estimated.
(v) Sampling to select the model to be trained in the next iteration based on the weighting.
The model to be updated is resampled with a probability proportional to the calculated weight value.
    }
    \label{fig:proposed_overview}
  \end{center}
\end{figure*}
NPB-DAA estimates latent phonemes, latent words, language models, and acoustic models, which are the latent variables of HDP-HLM, using the blocked Gibbs sampler\footnote{{HDP-HLM is the name of the probabilistic generative model, and NPB-DAA is the name of the inference algorithm for finding phonemes and words in HDP-HLM by blocked Gibbs sampler.}}.
{NPB-DAA uses a nonparametric Bayesian method (specifically, the stick-breaking process (SBP)~\cite{sethuraman1994constructive}, which is based on the Dirichlet process) to automatically estimate the appropriate number of categories (i.e., number of phonemes and word types) for the data. 
In practice, a weak-limit approximation~\cite{fox2011sticky} in the SBP is used to specify the maximum limit number of categories for implementation.}
For the inference algorithm, refer to the paper in which it was defined~\cite{taniguchi2016nonparametric}.

Figure~\ref{fig:hdphlm} shows the graphical model representation of HDP-HLM, which is a generative model.
The generative process is omitted in this paper.
In HDP-HLM, latent words continuously generate observation data for a certain period.
In addition, the latent word corresponds to the word $ z_s $, and the $ i $-th word $ z_s = i $ has the phoneme string $ w_i = (w_ {i1}, \cdots, w_ {ik}, \cdots, w_ {iL_ {i}}) $.
Here, $ L_i $ represents the number of phonemes of $ i $-th word $ w_i $.
The superscripts $ {\rm LM} $ and $ {\rm WM} $ represent the language model and the word model, respectively.
A word model is part of a language model that expresses the kind of phonemes each word is composed of and is referred to as a word dictionary.
$ \beta^{{\rm LM}} $ and $ \beta^{{\rm WM}} $ are base measures for the language model and the word model, respectively.
In addition, $ \alpha^{{\rm LM}} $, $ \gamma^{{\rm LM}} $, $ \alpha^{{\rm WM}} $, $ \gamma^{{\rm WM }} $ are the hyperparameters of the language model and the word model, respectively.
$ \pi^{{\rm LM}} _ {j} $ is the output from $ \rm{DP} ( \alpha^{{\rm LM}}, \beta^{{\rm LM}} )$, which expresses the transition probability.
$\pi^{{\rm WM}}_{j}$ is the output from $ \rm{DP} (\alpha^{{\rm WM}}, \beta^{{\rm WM}})$, which expresses the transition probability of the next latent character string from a latent phoneme $ j $.
$ w_ {ik} $ represents the $ k $-th latent phoneme in the $ i $-th latent word.
In addition, $ l_ {sk} $ is the $ k $-th latent character in the $ s $-th latent word $ z_s $.
$ \omega_ {l_ {sk}} $ is a parameter of the duration distribution of the latent character $ l_ {sk} $.
In HDP-HLM, the latent word $ z_s $ is generated by the previous latent word $ z_ {s-1} $ and the language model.
The duration $ D_ {sk} $ of $ l_ {sk} $ is sampled based on the determined sequence $ w_ {z_ {s}} $.
The observation data $ y_t $ is generated from the output distribution $ h (\theta_ {x_ {t}}) $ corresponding to $ x_t = l_ {s (t) k (t)} $.
Here, the map functions $ s (t) $ and $ k (t) $ represent the word and phoneme indicators of the latent word string at time $ t $, respectively.
Here, the observation time-series data $ y_t $ is associated with the feature vector obtained from the audio signal at time $ t $.

\section{Proposed Method: Unsupervised Phoneme and Word Discovery Method with Co-occurrence Cues} \label{sec:proposed_method}

In this section, we describe the proposed method, which performs unsupervised phoneme and word discovery using multimodal sensor data obtained by a robot.
{
The integration of the two methods involves one module sending inference results to the other module, and iterative learning improves overall learning.
Based on the information about the object category, word segmentation is more accurately corrected.
Using word segmentation results, better object categories are formed.
Initially, uncertain categories or incorrect words are gradually self-organized and corrected.
}

The proposed model, \textbf{HDP-HLM+MLDA}, is the integration of HDP-HLM (NPB-DAA) and MLDA {(Fig.~\ref{fig:int})}.
An overview of the proposed inference algorithm, \textbf{co-occurrence DAA}\footnote{{HDP-HLM+MLDA is the name of the probabilistic generative model, and co-occurrence DAA is the name of the inference algorithm in HDP-HLM+MLDA.}}, is shown in Fig.~\ref{fig:proposed_overview}.
The inference algorithm is realized by sampling importance resampling (SIR)~\cite{rubin1988using}, which samples candidates of word sequences using NPB-DAA and weights the candidates using the MLDA.
In other words, this algorithm performs iterative learning with NPB-DAA and MLDA.

\subsection{HDP-HLM+MLDA: Building an Integrated Probabilistic Generative Model} 
\label{sec:proposed_method:generative}

To integrate HDP-HLM and MLDA, we adopt the idea of the Symbol Emergence in Robotics Tool KIT (SERKET)~\cite{nakamura2018serket,taniguchi2020neuroSERKET}, which is an integration framework for probabilistic generative models.
SERKET makes it possible to easily construct a large-scale generative model and its inferences by hierarchically connecting the base models, which are its constituent units, while maintaining the independence of each program that is the integration source.
By constructing the integrated model according to the SERKET framework, it is possible to optimize the parameters of the integrated model, even if the parameters estimated independently in each base model are used.
In the proposed method, $ \Vec{o}^w $ corresponding to the word sequences is shared by HDP-HLM and MLDA.

\begin{figure}[tb]
  \begin{center}
    \includegraphics[width=\linewidth]{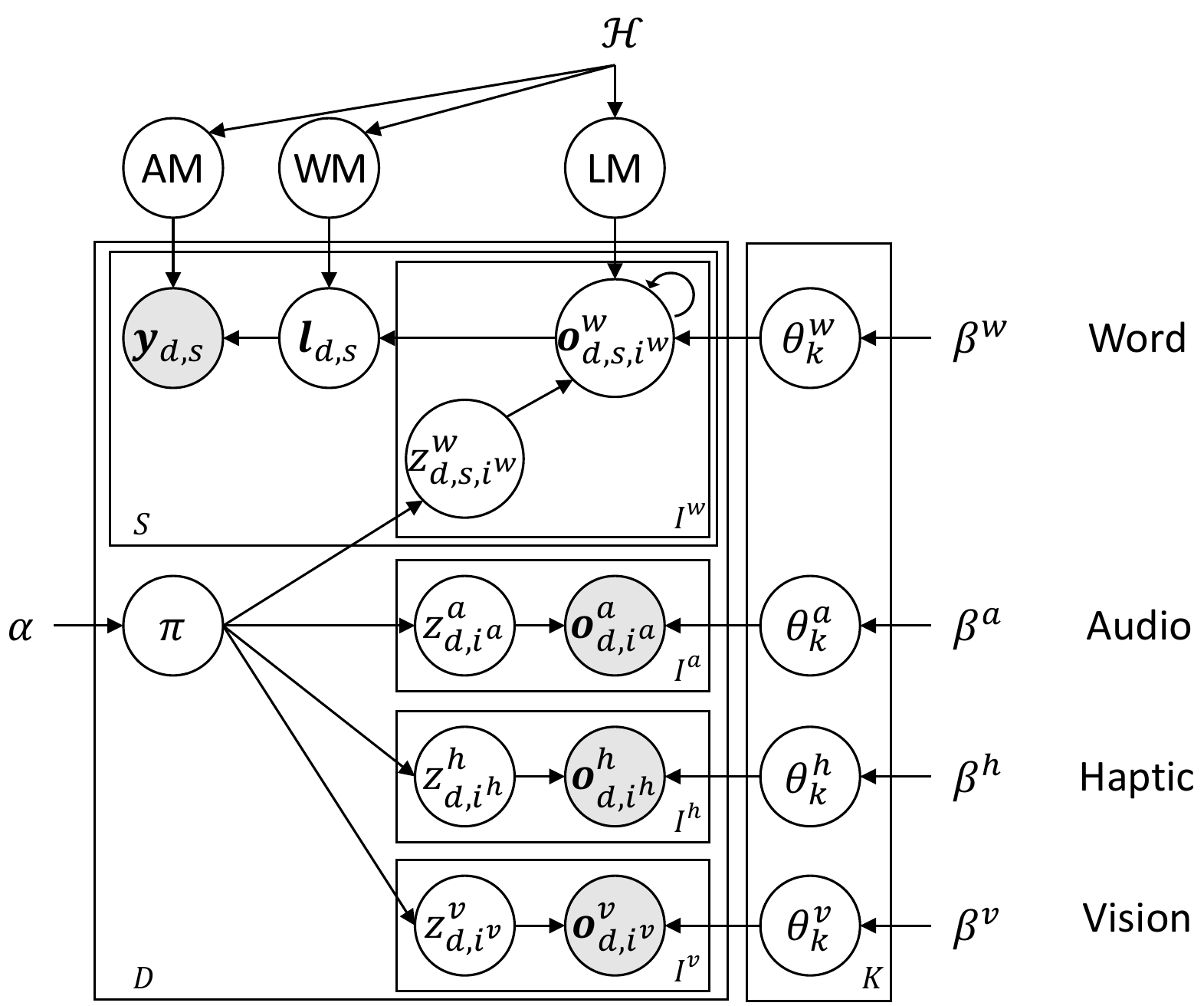}
    \caption{Graphical model representation of HDP-HLM+MLDA, which is the integrated model of HDP-HLM and MLDA.
    Some of the variables corresponding to HDP-HLM are collectively shown as one variable. 
    }
    \label{fig:int}
  \end{center}
\end{figure}

A graphical model of the proposed method is shown in Figure~\ref{fig:int}.
Each variable follows the definition in the graphical model of the base models shown in Section~\ref{sec:preparation}.
In the proposed graphical model, the part corresponding to HDP-HLM is expressed, and some variables are changed to avoid duplication.
Here, the word sequences $\Vec{o}^w$, the language model (LM), the word model (WM), including the word dictionary, and the acoustic model (AM) correspond to $\Vec{z_{s}}$, $\pi^{{\rm LM}}$, $\{ \pi^{{\rm WM}}, \Vec{W} \}$, $\{ \Vec{\omega}, \Vec{\theta} \}$ in the graphical model of HDP-HLM, respectively.

The probability distribution to generate a word sequence $P(\Vec{o}^w \mid \Vec{z}^w,\Vec{\theta}^w,\mathcal{G})$ can be defined using unigram rescaling (UR) approximation~\cite{gildea1999topic}.
The UR approximation represents category-dependent N-gram word probability as follows:
\begin{eqnarray}
    &&P(\Vec{o}^w \mid \Vec{z}^w,\Vec{\theta}^w,\mathcal{G}) \nonumber \\
    &&\stackrel{\mathrm{UR}}{\approx} \underbrace{P(\Vec{o}^w \mid \mathcal{G})}_{\text{N-gram prob.}} \prod_{d,s,i^w} \underbrace{\frac{P(\Vec{o}^w_{d,s,i^w} \mid  z^w_{d,s,i^w}=k,\theta^w_k)}{P(\Vec{o}^w_{d,s,i^w})}}_{\substack{\text{Category dependent term} \\ \text{/ Rescaling term}}},
    \label{eq:ur}
\end{eqnarray}        
where the global parameters of HDP-HLM related to the proposed method are denoted as $\mathcal{G}=\{\text{AM},\text{WM},\text{LM}\}$.

\subsection{Co-occurrence DAA: Procedure of the Inference Algorithm by NPB-DAA and MLDA}  \label{sec:proposed_method:inference}

The inference algorithm applies SIR to the UR approximation based on the SERKET framework~\cite{nakamura2018serket}.
The target distribution is the posterior category-dependent word probability distribution $P(\Vec{o}^w \mid \Vec{y},\Vec{z}^w,\Vec{\theta}^w,\mathcal{G})$.
The proposal distribution is the N-gram word probability distribution $P(\Vec{o}^w \mid \Vec{y}, \mathcal{G})$, which is estimated by NPB-DAA.
Resampling is performed according to the weights provided by the word distribution in the object category by MLDA.
This learning procedure enables the proposed method to acquire the lexicon considering the object categorization results by MLDA and to categorize objects using the word sequences estimated by NPB-DAA.

Specifically, the following procedures and formulas are used for learning.
First, the model parameters are initialized.
The initialization is the same as in {the previous NPB-DAA paper}~\cite{taniguchi2016nonparametric}.
The set of initial parameters $\mathcal{G}^{(q)}_{0}$ (as $t=0$) is sampled as $Q$ candidates independently.
Next, the following procedure, from I.--V., is iterated $T$ times ($t\in\{1,2,\dots,T\}$):
\begin{itemize}
    \item[I.]
        To generate word sequences with consideration of the object categories using SIR, we sample $Q$ candidates of the proposed distribution proportional to their respective weight values from the candidates sampled at the ($t-1$)-th iteration.
        Then, the parameters of each candidate are updated, and the word sequences are estimated as follows:
        \begin{eqnarray}
            \mathcal{G}^{(q)}_{\text{pre}} &\sim& \sum_{q=1}^{Q}{\mathcal{W}}\left( \mathcal{G}^{(q)}_{t-1} \right) \delta \left( {\mathcal{G}_{t-1}} - \mathcal{G}^{(q)}_{t-1} \right), \label{eq:MC_sampling} \\ 
            \Vec{o}^{w(q)}, \mathcal{G}^{(q)}_t &\sim& \text{NPB-DAA}\left( \Vec{y},\mathcal{G}^{(q)}_{\text{pre}},\mathcal{H} \right), 
        \end{eqnarray}
        where the set of hyperparameters of HDP-HLM is $\mathcal{H}=\{{\rm G},{\rm H},\gamma^{\rm LM},\alpha^{\rm LM},\gamma^{\rm WM},\alpha^{\rm WM}\}$, 
        the $q$-th global parameter candidate of HDP-HLM at $t$-th iteration is $\mathcal{G}^{(q)}_{t}$, and
        the global parameter candidate resampled by weight ${\mathcal{W}}\left( \mathcal{G}^{(q)}_{t-1} \right)$ at ($t-1$)-th iteration is $\mathcal{G}^{(q)}_{\text{pre}}$.
        In addition, $\delta\left( \cdot \right)$ is the Dirac delta mass in Eq.~(\ref{eq:MC_sampling}).
        Note that in the ($t=1$)-th iteration, the weight in the ($t-1$)-th iteration does not exist, so the initial value is copied as $\mathcal{G}^{(q)}_{0} = \mathcal{G}^{(q)}_{\text{pre}}$.
        Here, $\text{NPB-DAA}(\cdot)$ is the process of one iteration of the blocked Gibbs sampler by NPB-DAA.
    \item[II.]
        Object categorization using MLDA uses $ q $-th word sequences candidate ${\Vec{o}}^{w(q)}$ and co-occurrence cues $\Vec{o}^{a,h,v}$. 
        The global parameters $\theta^{w(q)}$ and $\pi^{(q)}$ are obtained as follows:
        \begin{eqnarray}
            \theta^{w(q)}, \pi^{(q)} \sim \text{MLDA}\left( \Vec{o}, \alpha, \beta \right).
            \label{eq:MLDA_sampling}
        \end{eqnarray}
        Here, the set of word sequences for each candidate $\Vec{o}^{w(q)}$ is converted to a bag-of-words representation.
        We apply the collapsed Gibbs sampler until the MLDA categorization is sufficiently converged.
        The above process is performed for each set of word sequences ${\Vec{o}}^{w(q)}$ estimated by all $ Q $ parameter candidates.
    \item[III.]
        The category $ \hat{k} $ assigned to each word is sampled from the probability distribution based on $\theta^{w(q)}$ and $\pi^{(q)}$
        \footnote{
        It is also possible to refer to the category index $\Vec{z}^{w}$ assigned to a word directly from the categorization process by MLDA.
        However, owing to the code implementation used, we decided to perform sampling again in this study.
        }.
        The probability that the category $ k $ is assigned to each word $o^{w(q)}_{d,s,i^w}$ is as follows:
        \begin{eqnarray}
            \hat{k} &\sim& P\left(z^{w(q)}_{d,s,i^w}=k\mid o^{w(q)}_{d,s,i^w},\pi^{(q)}_{d,k},\theta^w_k\right) \nonumber \\
            &&= \frac{\theta^{w(q)}_{k,o^{w(q)}_{d,s,i^w}} \pi^{(q)}_{d,k}}{\sum_{k}\left(\theta^{w(q)}_{k,o^{w(q)}_{d,s,i^w}} \pi^{(q)}_{d,k}\right)}.
            \label{eq:Pz_dsi}
        \end{eqnarray}    
    \item[IV.]
        The weight of the parameter candidate of each HDP-HLM, corresponding to the second term on the right side in Eq.~(\ref{eq:ur}) is calculated as follows:
        \begin{eqnarray}
            \overline{\mathcal{W}}\left(\mathcal{G}^{(q)}_t\right) &=& \prod_{d,s,i^w} \frac{P\left( o^{w(q)}_{d,s,i^w}\mid z^{w(q)}_{d,s,i^w}=\hat{k},\theta^w \right)}{P\left( o^{w(q)}_{d,s,i^w} \right)} \nonumber\\
            &=& \prod_{d,s,i^w}\frac{\theta^{w(q)}_{\hat{k},o^{w(q)}_{d,s,i^w}}}{\sum_{k}\theta^{w(q)}_{k,o^{w(q)}_{d,s,i^w}}}.
        \end{eqnarray}
        The weight is then normalized to use probability:
        \begin{eqnarray}
            {\mathcal{W}}\left(\mathcal{G}^{(q)}_t\right) = 
            \frac{\overline{\mathcal{W}}\left(\mathcal{G}^{(q)}_t\right)}{\sum_q\overline{\mathcal{W}}\left(\mathcal{G}^{(q)}_t\right)}.
        \end{eqnarray}
        This normalized weight is held until the ($t+1$)-th iteration and used to sample candidates.
    \item[V.]
        The candidate with the largest weight is adopted as the estimation result in the $t$-th iteration.
        Increase iteration value ($t\leftarrow t+1$) and return to Step I.
\end{itemize}

\subsection{How to Weight Each Modality for Multimodal Observations}
\label{sec:proposed_method:modality_weight}

Similar to MLDA, the proposed method introduces weighting to adjust the degree of influence on categorization by modality.
Weighting in MLDA increases the quantity of the feature itself (i.e., increasing the frequency of each occurrence of the histogram).
By changing the word modality weight, it is possible to investigate the effect on categorization in lexical acquisition.

The weighting process of categorization for each modality is calculated as follows:
\begin{eqnarray}
    \Vec{o}^{m} = \frac{\texttt{hist}^{m}}{\sum \texttt{hist}^{m}} \times \texttt{modality\_weight}^{m},
\end{eqnarray}
where $\texttt{hist}^{m}$ is the original observation feature histogram and $\texttt{modality\_weight}^{m}$ is the weight value of a modality $m$.

\section{Spoken Utterance and Multimodal Dataset} \label{sec:dataset}
This section describes the dataset of spoken utterances for phoneme and word discovery.

\subsection{Overview}

\begin{figure}[!t]
  \begin{center}
    \includegraphics[width=\linewidth]{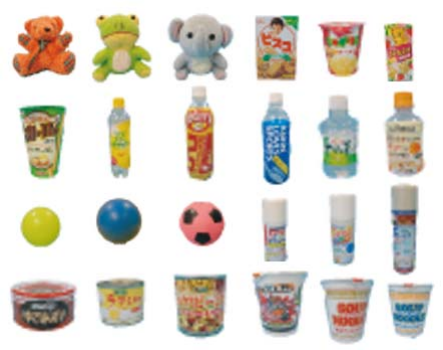}
    \caption{List of objects used in the experiment from Multimodal Object Dataset 165~\cite{nakamura2017ensemble}.}
    \label{fig:obj}
  \end{center}
\end{figure}

\begin{table}[tb]
    \centering
    \caption{Examples of uttered sentences.}    
    \begin{tabular}{cc}\hline
    \textbf{Uttered sentences (Japanese phoneme)} & \textbf{English} \\ \hline 
    /kore wa omocha/ & This is a toy. \\ 
    /jyuusu no botoru dayo/ & It's a bottle of juice. \\ 
    /supurei kaN wa katai/ & The spray can is hard. \\ \hline
    \end{tabular}
    \label{tab:hatsuwa}
\end{table}

To evaluate the performance of speech unit discovery with co-occurrence cues, we generated a speech dataset corresponding to the multimodal object sensor data.
We used the Multimodal Object Dataset 165~\cite{nakamura2017ensemble}, which is an open dataset that includes vision, haptic, and audio sensory data, as well as multimodal co-occurrence cues\footnote{Multimodal Object Dataset 165:\\ \url{http://hp.naka-lab.org/subpages/mod165.html}}. 
Here, we used 24 objects from the dataset for experiments.
See {Nakamura et al.'s paper}~\cite{nakamura2017ensemble} for details regarding the observation process for each modality.

Figure~\ref{fig:obj} shows the image list for objects in the dataset.
Objects were categorized into one of seven potential categories: stuffed toys, sweets, bottles, balls, spray cans, food cans, or cup noodles.
Table~\ref{tab:hatsuwa} shows an example of speech sentences, with the Japanese phonemes and the English translation.
The speech dataset is the content that teaches the characteristics and names of each object for a total of 75 Japanese sentences.
{Each speech item had a duration of approximately 2--3 seconds per utterance.}

\subsection{Procedure for Creating Speech Dataset}

A Japanese speaker was recorded in an anechoic chamber using an omnidirectional microphone (SHURE PG27-USB).
The speech was uttered as clearly as possible to avoid speech recognition errors.
The speech was saved as a 1-channel wav file with a sampling frequency of 16.1 kHz.
The silence intervals before and after the utterance were removed using the automatic speech recognition system Julius\footnote{Open-Source Large Vocabulary Continuous Speech Recognition Engine Julius: \url{https://github.com/julius-speech/dictation-kit}}~\cite{lee2001julius} because they would greatly affect the accuracy of phoneme and word discovery if they remained in the dataset.
Next, the Mel-frequency cepstral coefficients (MFCC) and the first and second derivatives of the MFCC were extracted from the speech data as features.
The MFCC features were extracted with the frame width set to 25 msec and the frameshift length set to 10 msec.
The MFCC and first and second derivatives of the MFCC were 12-dimensional features.
In this study, we used a deep sparse auto-encoder with parametric bias in the hidden layer (DSAE-PBHL)~\cite{nakashima2019unsupervised} to extract 12D, 8D, 5D, and 3D features in a stepwise manner. 
Here, the features were compressed using DSAE-PBHL because (i) the experimental results of a previous study~\cite{nakashima2019unsupervised} showed higher word discovery accuracy compared to MFCC, and (ii) dimensional reduction reduces the computational cost.
Using the above procedure, each speech utterance was used as a 9-dimensional acoustic feature.

\section{Experiment~1: Phoneme and Word Discovery using Co-occurrence Cues Obtained by Observing Real Objects} \label{sec:ex_1}

In this experiment, we compared the performance of the proposed method, which uses speech and its co-occurrence cues, with that of NPB-DAA, which uses only speech signals.
We investigated {the hypothesis that} the use of co-occurrence information contributes to the improvement of phoneme and word discovery accuracy in situations close to real-world environments.
We also investigated whether the word sequences discovered by exploiting co-occurrence with object information also affect the performance of object categorization.
The co-occurrence information is described in Section~\ref{sec:dataset}, and the experiments were conducted using a multimodal object dataset.

\subsection{Condition} \label{sec:ex_1:condition}
This experiment used the dataset described in Section~\ref{sec:dataset}.
The hyperparameters of the language model LM of HDP-HLM were $\alpha^{\text{LM}}=10.0$ and $\gamma^{\text{LM}}=10.0$.
The limit of the number of words by weak-limit approximation was 50 words.
The hyperparameters of the word model WM of HDP-HLM were $\alpha^{\text{WM}}=10.0$ and $\gamma^{\text{WM}}=10.0$.
The limit of the number of phonemes by weak-limit approximation was 50 phonemes.
The duration distribution assumed a Poisson distribution of $\alpha_0=200$ and $\beta_0=10$.
The emission distribution of the acoustic features assumed a multivariate Gaussian distribution.
The prior distribution is a normal inverse Wishart distribution of $\mu_0=0$, $\Sigma_0=I$ (unit matrix), $\kappa_0=0.01$, and $\nu_0=14$ $(=\text{Dimension}+5=9+5)$.
Here, we used \texttt{NPB\_DAA}\footnote{NPB-DAA: \url{https://github.com/EmergentSystemLabStudent/NPB\_DAA}} as the source code for the NPB-DAA implementation.
We also set the number of candidate parameters for HDP-HLM in each iteration to $Q=10$.

MLDA uses the histograms of the four modalities as input, and the number of object categories to be set in the prior was $K=7$.
The hyperparameter of the Dirichlet prior distribution, which is prior for multinomial distribution representing the probability of appearance of each category, was set to $\alpha=7.1$ $(\approx50/K=50/7)$.
The hyperparameter of the Dirichlet prior distribution, which represents the emission probability of features in each category, was set to $\beta=0.1$. 
Here, we used \texttt{LightMLDA}\footnote{MLDA: \url{https://github.com/naka-tomo/LightMLDA}} as the source code for the MLDA implementation.
The weight values set by MLDA for each modality were fixed.
The details are described in Section~\ref{sec:ex_1:method}.
Under the above conditions, one trial consisted of 100 iterations of a blocked Gibbs sampler of the NPB-DAA, and 20 trials were performed independently.
In addition, MLDA takes 1000 iterations of a Gibbs sampler for each candidate of the word sequences for each iteration of NPB-DAA.

\subsection{Comparison Methods}\label{sec:ex_1:method}

The comparison methods are described as follows:
    \\
    \textbf{NPB-DAA}~\cite{taniguchi2016nonparametric}:
    This method is the conventional baseline method for phoneme and word discovery with only distributional cues.
    \\
    \textbf{MLDA}~\cite{nakamura2011grounding}:
    This is the conventional method for object categorization with multimodal data.
    To evaluate categorization performance, two types were performed using (i) only multimodal data, that is, 3 modalities (audio, haptic, and vision) as the baseline method, and (ii) transcription sentences (ground truth) with multimodal data, that is, four modalities (word, audio, haptic, and vision) as the topline method.
    \\
    \textbf{HDP-HSMM}~\cite{johnson2013}\textbf{ + MLDA}:
    This is a method that integrates the hierarchical Dirichlet process hidden semi-Markov model (HDP-HSMM) and MLDA within the framework of SERKET, which is similar to the proposed method.
    We created this baseline to demonstrate the advantages of using co-occurrence in DAA.
    HDP-HSMM is used instead of HDP-HLM to use the distributional cues.
    HDP-HSMM does not assume double articulation for phonemes and words.
    Therefore, during evaluation, the segmented results are applied to both phonemes and words.
    \\
    \textbf{Co-occurrence DAA (Proposed method)}:
    The modality weights ($\texttt{modality\_weight}^{m}$) for object categorization (See Section~\ref{sec:proposed_method:modality_weight}) are $\texttt{word}:\texttt{audio}:\texttt{haptic}:\texttt{vision} = 200:50:100:100$ as Co-occurrence DAA$^{\ast 1}$ and $\texttt{word}:\texttt{audio}:\texttt{haptic}:\texttt{vision} = 200:340:160:280$ as Co-occurrence DAA$^{\ast 2}$.
    The $\texttt{modality\_weight}^{m}$ of the method ${\ast 1}$ is determined by preliminary experiments (See Appendix~\ref{sec:apdx}).
    The $\texttt{modality\_weight}^{m}$ of the method ${\ast 2}$ is used, implementing the values established by a previous study~\cite{nakamura2014mutual}.
    \\
    \textbf{Julius GMM/DNN~\cite{lee2001julius}}:
    This is the topline, a speech recognizer built by supervised learning from large labeled speech datasets.
    Julius GMM uses a Gaussian mixture model-based triphone model, and Julius DNN uses a deep-neural-network-based triphone model.
    The two different types of word dictionaries are prepared; 
    a generic word dictionary is a built-in large-scale word dictionary, and a true word dictionary consists only of the words in the dataset.
    These results were taken as reference values from Experiment 2 in {Okuda et al.'s paper}~\cite{okuda2021prosody}, where a similar dataset is used. 

\subsection{Evaluation Metrics} \label{sec:ex_1:eval}

We prepared latent letters, that is, phonemes, and word ground truth labels for all datasets and evaluated the relationship between the ground truth labels and estimated latent letters and words as word discovery performance. 
We used the automatic annotation tool provided by Julius GMM to prepare the ground truth labels.
We also evaluated the relationship between the ground truth of object categorization by the tutor and the estimated object categorization results as categorization performance.

The evaluation metrics were as follows:
\\
    \textit{Normalized Mutual Information~(NMI)}~\cite{kvalseth1987entropy}:
    NMI is one of the most widely used evaluation metrics in clustering tasks for unsupervised learning.
    NMI is an evaluation value obtained by normalizing the amount of mutual information between the correct clustering result and the estimated clustering result to take a value ranging from 0.0 to 1.0.
    NMI is evaluated for phoneme, word, and object categories.
    \\
    \textit{Adjusted Rand Index~(ARI)}~\cite{hubert1985comparing}:
    ARI is one of the most widely used evaluation metrics in clustering tasks in unsupervised learning.
    ARI takes 1.0 when the clustering result matches the correct label and 0.0 when it is random.
    ARI is evaluated for phoneme, word, and object categories.
    \\
    \textit{Object categorization accuracy~(ACC)}:
    ACC is a metric used to evaluate the performance of object categorization in a series of studies on MLDA~\cite{nakamura2014mutual, nakamura2011grounding}.
    This metric represents the matching rate when the label is changed so that the estimated clustering label value most closely matches the correct clustering label value.

\subsection{Results} \label{sec:ex_1:result}

\begin{table*}[!tb]
    \centering
    \caption{ 
    Phoneme and word discovery performance. (Experiment~1)
    }
    \begin{tabular}{ccccccc} \hline
        \textbf{Methods} & \textbf{Distributional cue} & \textbf{Co-occurrence cue} & \textbf{Phoneme NMI} & \textbf{Word NMI} & \textbf{Phoneme ARI}& \textbf{Word ARI} \\ \hline 
        NPB-DAA                      & \checkmark &            & $0.556\pm0.008$ & $0.722\pm0.023$ & $0.307\pm0.017$ & $0.519\pm0.049$ \\ 
        HDP-HSMM+MLDA                & w/o DAA    & \checkmark & $0.550\pm0.008$ & $0.412\pm0.015$ & $0.292\pm0.017$ & $0.167\pm0.011$ \\ 
        Co-occurrence DAA$^{\ast 1}$ & \checkmark & \checkmark & $0.556\pm0.010$ & {$\mathbf{0.731\pm0.028}$} & $0.307\pm0.017$ & {$\mathbf{0.548\pm0.056}$} \\ 
        Co-occurrence DAA$^{\ast 2}$ & \checkmark & \checkmark & $0.557\pm0.007$ & \underline{$\mathbf{0.751\pm0.020}$} & $0.311\pm0.013$ & \underline{$\mathbf{0.575\pm0.050}$} \\ \hline
        {Julius GMM with generic word dict.} &            &            & --- & --- & {$0.575$} & {$0.557$} \\ 
        {Julius DNN with generic word dict.} &            &            & --- & --- & {$0.474$} & {$0.725$} \\ \hline
        {Julius GMM with true word dict.}    &            &            & --- & --- & {$0.677$} & {$0.900$} \\ 
        {Julius DNN with true word dict.}    &            &            & --- & --- & {$0.493$} & {$0.825$} \\ \hline
    \end{tabular}
    \label{tab:ex_1_last_ARI}
\end{table*}

\begin{table*}[!tb]
    \centering
    \caption{Object categorization performance; accuracy (ACC), NMI, and ARI. (Experiment~1)}    
    \begin{tabular}{ccccc}\hline
    \textbf{Methods} & \textbf{Word modality} & \textbf{ACC} & \textbf{NMI} & \textbf{ARI} \\ \hline 
    MLDA                         & Ground truth         & $0.875\pm0.000$ & $0.862\pm0.000$ & $0.725\pm0.000$\\ \hline    
    MLDA                         & No use               & $0.563\pm0.000$ & $0.554\pm0.000$ & $0.205\pm0.000$\\ 
    HDP-HSMM+MLDA                & Iterative estimation & $0.550\pm0.096$ & $0.619\pm0.059$ & $0.249\pm0.103$\\ 
    Co-occurrence DAA$^{\ast 1}$ & Iterative estimation & $\mathbf{0.677\pm0.084}$ & $\mathbf{0.730\pm0.078}$ & $\mathbf{0.438\pm0.135}$\\ 
    Co-occurrence DAA$^{\ast 2}$ & Iterative estimation & \underline{$\mathbf{0.700\pm0.069}$} & \underline{$\mathbf{0.772\pm0.053}$} & \underline{$\mathbf{0.517\pm0.112}$}\\ \hline
    \end{tabular}
    \label{tab:ex_1_cat_ARI}
\end{table*}

Table~\ref{tab:ex_1_last_ARI} shows the evaluation results for phonemes and words at the end of training.
The proposed method has a higher word discovery performance than the baseline methods.
As a result, we have shown that using co-occurrence cues improves word discovery performance in lexical acquisition.
In contrast, the phoneme discovery performance was almost the same for all methods.
This is likely because phonemes are the smallest speech units without meaning, whereas words are the speech units that can be assigned meaning.

Table~\ref{tab:ex_1_cat_ARI} shows the performance of object categorization at the end of training.
The proposed method showed better categorization performance than MLDA.
The proposed method showed higher values with the modality parameters that were empirically set by {Nakamura et al.}~\cite{nakamura2014mutual} than with the modality parameters determined by preliminary experiments.
HDP-HSSM+MLDA performed poorly in speech segmentation and inaccurately in categorization because it did not assume double articulation.
As a result, more accurate word discovery resulted in higher categorization performance.
The results also suggest that more accurate object categorization leads to higher word discovery performance.

\begin{figure*}[tb]
  \begin{subfigure}[b]{0.33\linewidth}
    \centering
    \includegraphics[width=\linewidth]{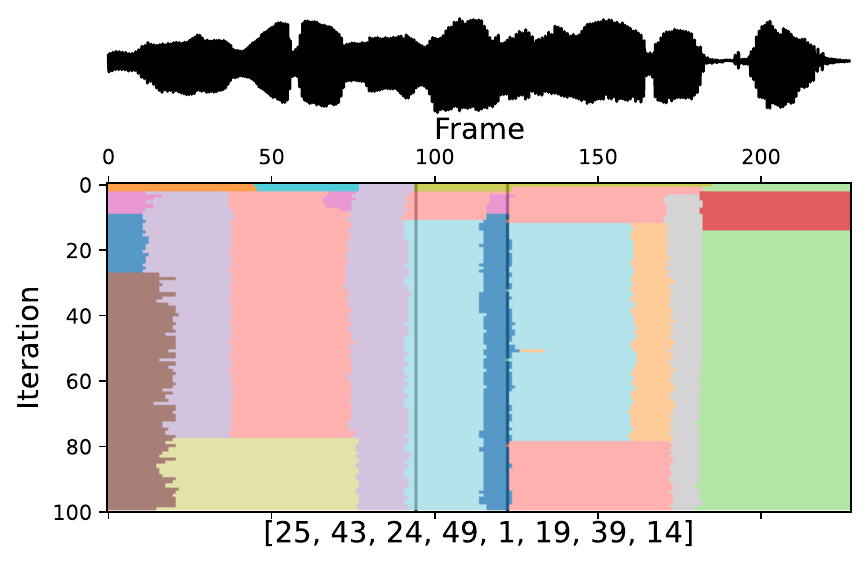} 
    \caption{NPB-DAA: \\ /\underline{nuigurumi} wa \underline{yawarakai}/}\footnotesize{(Plush is soft.)} 
    \label{fig:sub_1_segment1}
  \end{subfigure}
  \begin{subfigure}[b]{0.33\linewidth}
    \centering
    \includegraphics[width=\linewidth]{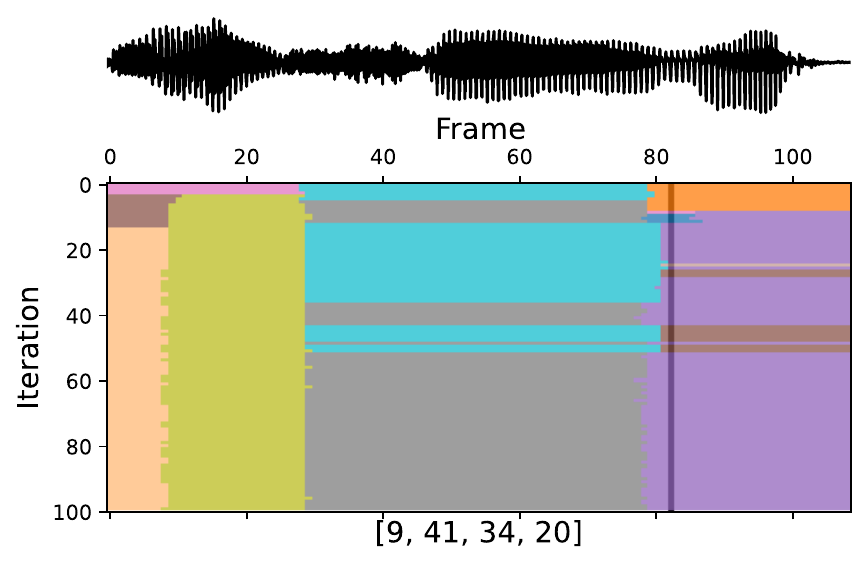} 
    \caption{NPB-DAA: \\ /\underline{oishii} ne/}\footnotesize{(It's delicious.)}
    \label{fig:sub_1_segment2}
  \end{subfigure}
  \begin{subfigure}[b]{0.33\linewidth}
    \centering
    \includegraphics[width=\linewidth]{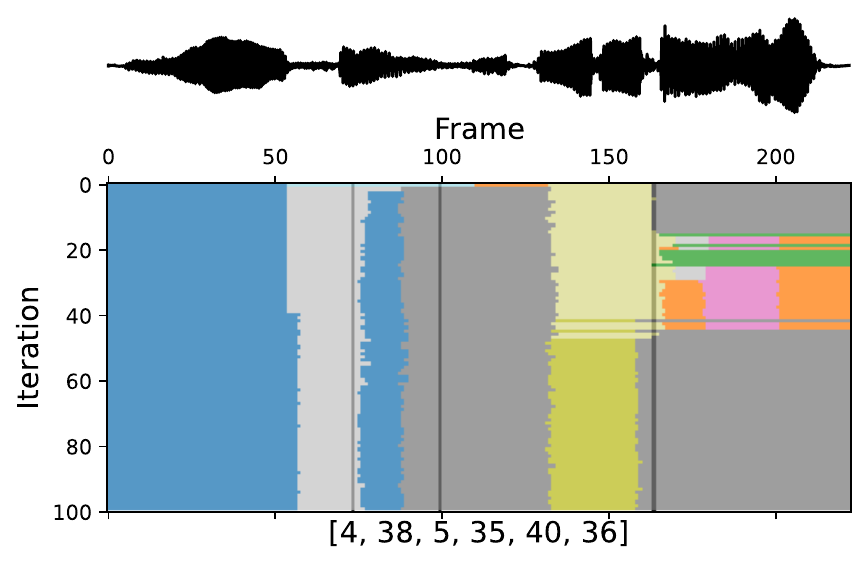} 
    \caption{NPB-DAA: \\ /\underline{ju:su} no \underline{botoru} dayo/}\footnotesize{(It's a bottle of juice.)} 
    \label{fig:sub_1_segment3}
  \end{subfigure}
  \\\\
    \begin{subfigure}[b]{0.33\linewidth}
    \centering
    \includegraphics[width=\linewidth]{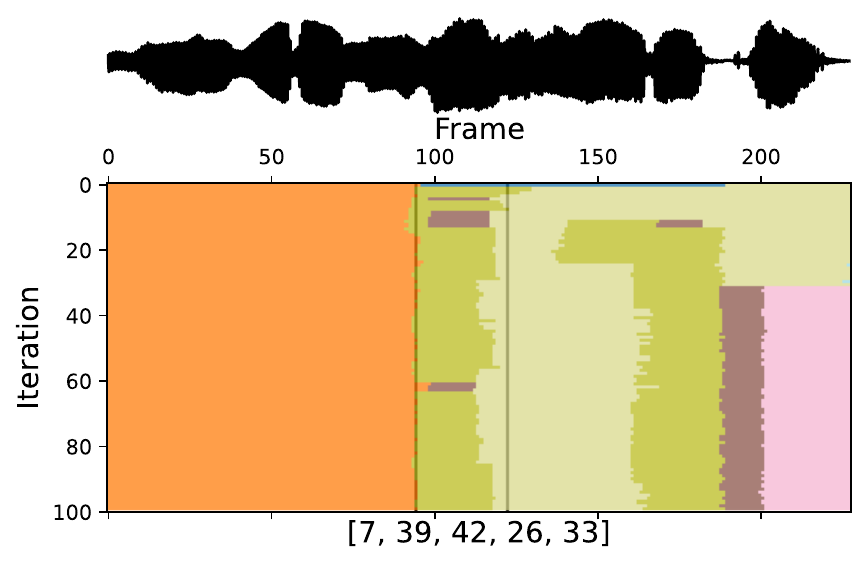} 
    \subcaption{Co-occurrence DAA: \\ /\underline{nuigurumi} wa \underline{yawarakai}/}\footnotesize{(Plush is soft.)} 
    \label{fig:ex_1_segment1}
  \end{subfigure}
  \begin{subfigure}[b]{0.33\linewidth}
    \centering
    \includegraphics[width=\linewidth]{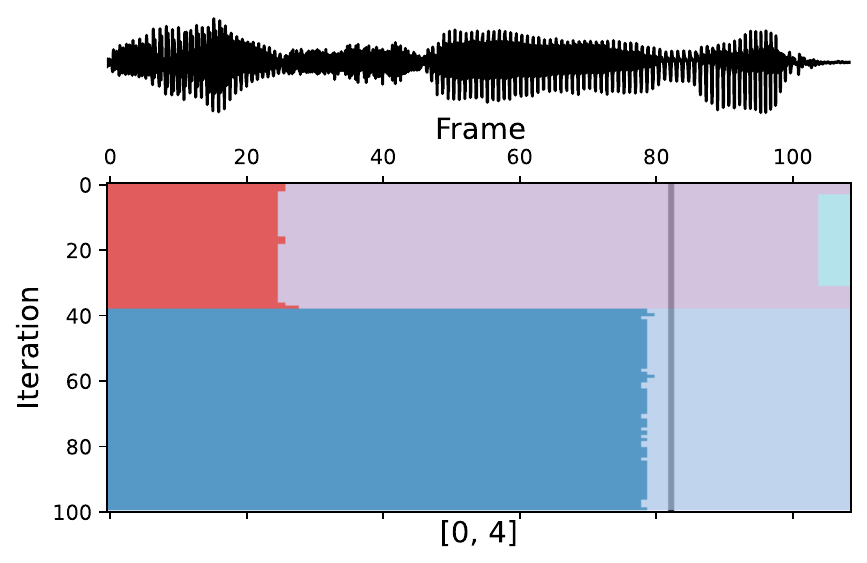} 
    \subcaption{Co-occurrence DAA: \\ /\underline{oishii} ne/}\footnotesize{(It's delicious.)}
    \label{fig:ex_1_segment2}
  \end{subfigure}
  \begin{subfigure}[b]{0.33\linewidth}
    \centering
    \includegraphics[width=\linewidth]{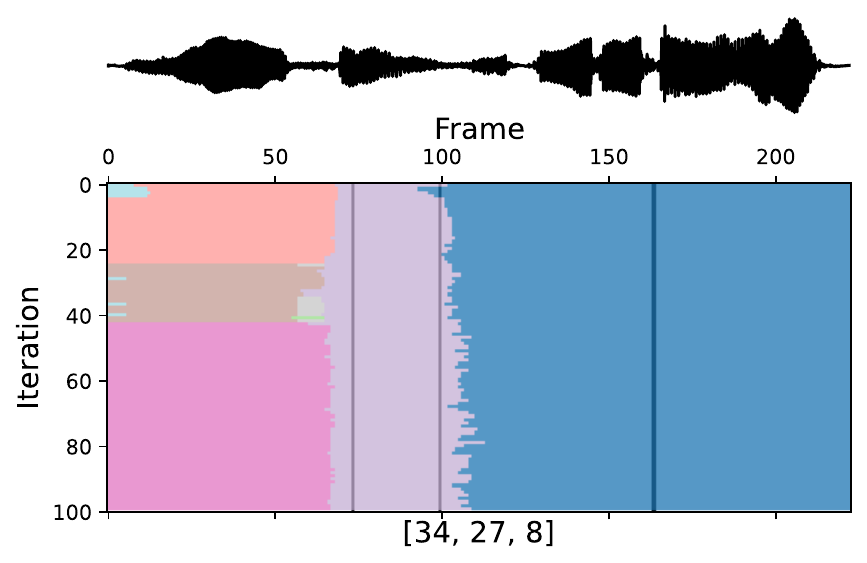} 
    \subcaption{Co-occurrence DAA: \\ /\underline{ju:su} no \underline{botoru} dayo/}\footnotesize{(It's a bottle of juice.)} 
    \label{fig:ex_1_segment3}
  \end{subfigure}
  \caption{Examples of word segmentation results (Experiment 1): 
The upper part of each sub-figure shows the waveform of the target speech.
The lower part of each sub-figure shows the segmentation position estimated by learning, color-coded for each word.
The correct segmentation position of the word is overlaid as a gray line.
The horizontal axis represents the number of speech frames and the vertical axis represents the training iteration. 
The list of numbers at the bottom of the lower part of each sub-figure corresponds to the index sequence of the words estimated by training.
The sub-caption shows the actual phoneme sequence of the utterance.
The underlined part is a characteristic word for an object.
  }
  \label{fig:segment1}
\end{figure*}

Figure~\ref{fig:segment1} shows examples of the results of word segmentation of speech.
These figures were drawn from the results at the last iteration of the trial with the highest word ARI.
Here, words that describe the characteristics of an object are nouns and adjectives such as ``sweets'' and ``soft.''
Most of the word segmentation results of NPB-DAA resulted in over-segmentation.
The word segmentation results of the proposed method reduce over-segmentation for words that describe the object characteristics.
For example, the word /nuigurumi/ is segmented as multiple words in Figure~\ref{fig:sub_1_segment1}, while it is correctly segmented as a single word in Figure~\ref{fig:ex_1_segment1}.
Additionally, Figure~\ref{fig:ex_1_segment2} is an example of almost exact word segmentation.
The proposed method correctly segmented the word /okashi/, a feature of the object in this utterance.
Figure~\ref{fig:ex_1_segment3} suppressed the over-segmentation of the word /ju:su/, but the words /botoru/ and /dayo/ were under-segmented.
As a result, the existing methods tended to consider a certain percentage of utterances as several words, whereas the proposed method could segment words that represented object features more accurately.

\section{Experiment~2: Effects of the Weight of Word Modality on the Performance of Word Discovery Object Categorization} \label{sec:ex_2}

By varying the word modality weights, we aimed to investigate the implications for lexical acquisition and object categorization.
{The hypothesis tested in this experiment is ``Uncertain word segmentation results in the early stage of learning can have a negative impact on classification and hinder overall performance improvement.''}

The proposed method is realized by the coupling of two modules and their mutual iteration. The weight of a modality in object categorization controls the degree of influence, that is, the importance of the modality when merging the two modules.
Therefore, adjusting the weight of a word modality may affect its classification performance.
Word discovery performance can also be affected by categorization because word discovery is acquired by exploiting co-occurrences with object information.
Therefore, in this experiment, we focused on the importance of word modality $\texttt{modality\_weight}^{w}$ described in Section~\ref{sec:proposed_method:modality_weight}.

As an additional evaluation for comparison, we performed a method using mutual information (MI) \cite{taniguchi2018unsupervised} instead of weighting based on unigram rescaling (UR).
The weighting by MI is equivalent to the logarithm of the weighting by UR.
MI provides a softer resampling of candidates than UR.

\subsection{Condition} \label{sec:ex_2:condition}

\begin{figure}[tb]
  \begin{center}
    \includegraphics[width=0.8\linewidth]{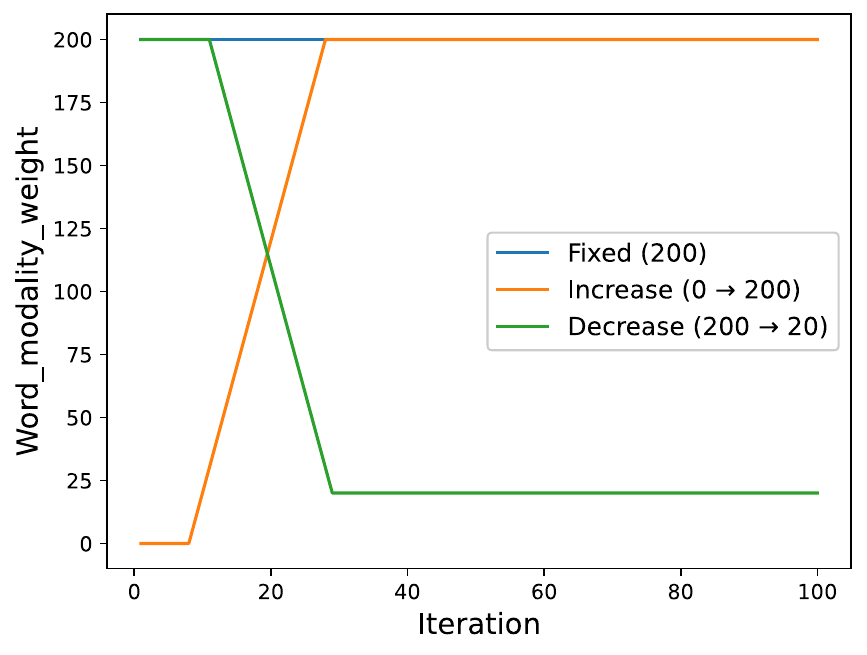}
    \caption{Visualization of conditions of word modality weight.}
    \label{fig:weight_set}
  \end{center}
\end{figure}

The dataset and hyperparameter settings were the same as those described in Section~\ref{sec:ex_1}.
For the weight settings of the word modalities, we applied variable weight settings in addition to the fixed value (200) used in the experiments in Section~\ref{sec:ex_1}.

\textbf{In the increase condition}, this weighting did not use the uncertain word segmentation results for categorization in the initial stage of learning and increased the weights after some progress in the NPB-DAA iteration.
This means that lexical acquisition and concept formation takes place through separate mechanisms in the early stage, followed by the integration of knowledge from both.
The increase condition of the word modality weight ($\texttt{modality\_weight}^{w}(t)$) was set according to the number $t$ of iterations of the blocked Gibbs sampler of NPB-DAA, as follows:
\begin{align}
  &\texttt{word\_modality\_weight\_Increase}(t)\nonumber \\
  &\quad  = 
  \max(0, {\min}(30 + 10 (t-10), 200) ).
  \label{eq:weight_set}
\end{align}

\textbf{In the decrease condition}, we set a change in weight that is considered inappropriate for comparison. 
This weighting strongly uses the word segmentation results for categorization in the early stages of learning but gives less credence to the word segmentation results.
The decrease condition of the word modality weight ($\texttt{modality\_weight}^{w}(t)$) was set according to the number $t$ of iterations of the blocked Gibbs sampler of NPB-DAA, as follows:
\begin{align}
  &\texttt{word\_modality\_weight\_Decrease}(t)\nonumber \\
  &\quad  = 
  \min( {\max}(20, 10 (30-t)), 200) ).
  \label{eq:weight_set_decrease}
\end{align}

Graphical plots of these conditions are shown in Fig~\ref{fig:weight_set}.
The above setting was determined based on Appendix~\ref{sec:apdx}.

\subsection{Results} \label{sec:ex_2:result}

\begin{table*}[tb]
    \centering
    \caption{
    Phoneme and word discovery performance and object categorization performance. (Experiment~2)
    } 
    \begin{tabular}{cccccccc} \hline
        \textbf{Methods} & \textbf{SIR} & \textbf{Word modality weight}& \textbf{Word NMI}& \textbf{Word ARI} & \textbf{Cat. ACC}  & \textbf{Cat. NMI} & \textbf{Cat. ARI} \\ \hline 
        $\ast 1$ & UR & Fixed (200)                    & $0.731\pm0.028$ & $0.548\pm0.056$ & $0.677\pm0.084$ & $0.730\pm0.078$ & $0.438\pm0.135$ \\
        $\ast 1$ & UR & Increase (0 $\rightarrow$ 200) & $0.763\pm0.030$ & $\mathbf{0.587\pm0.066}$ & $0.710\pm0.065$ & $0.765\pm0.046$ & $0.491\pm0.095$ \\
        $\ast 2$ & UR & Fixed (200)                    & $0.751\pm0.020$ & $0.575\pm0.050$ & $0.700\pm0.069$ & $0.772\pm0.053$ & $0.517\pm0.112$ \\
        $\ast 2$ & UR & Increase (0 $\rightarrow$ 200) & $\mathbf{0.764\pm0.026}$ & \underline{$\mathbf{0.595\pm0.055}$} & \underline{$\mathbf{0.742\pm0.086}$} & \underline{$\mathbf{0.782\pm0.083}$} & \underline{$\mathbf{0.538\pm0.154}$} \\
        $\ast 2$ & UR & Decrease (200 $\rightarrow$ 20)& $0.747\pm0.031$ & $0.563\pm0.070$ & $0.562\pm0.065$ & $0.620\pm0.051$ & $0.237\pm0.085$ \\
        $\ast 2$ & MI & Fixed (200)                    & $0.758\pm0.026$ & $0.581\pm0.063$ & $\mathbf{0.733\pm0.084}$ & $\mathbf{0.776\pm0.067}$ & $\mathbf{0.526\pm0.130}$ \\
        $\ast 2$ & MI & Increase (0 $\rightarrow$ 200) & \underline{$\mathbf{0.765\pm0.028}$} & $0.580\pm0.065$ & $0.719\pm0.068$ & $0.773\pm0.065$ & $0.521\pm0.102$ \\ \hline
    \end{tabular}
    \label{tab:ex_2_last_ARI}
\end{table*}

\begin{figure*}[tb]
  \begin{subfigure}[b]{0.33\linewidth}
    \centering
    \includegraphics[width=\linewidth]{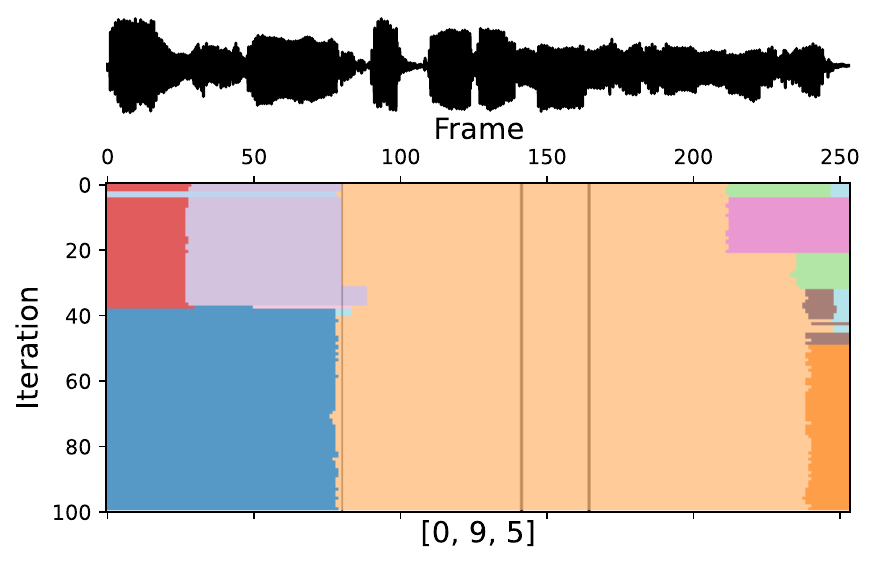} 
    \subcaption{Fixed condition: \\ /\underline{oishii} \underline{botoru} no \underline{nomimono}/ }
    \footnotesize{(A bottle of delicious drink.)} 
    \label{fig:ex_2_segment1}
  \end{subfigure}
  \begin{subfigure}[b]{0.33\linewidth}
    \centering
    \includegraphics[width=\linewidth]{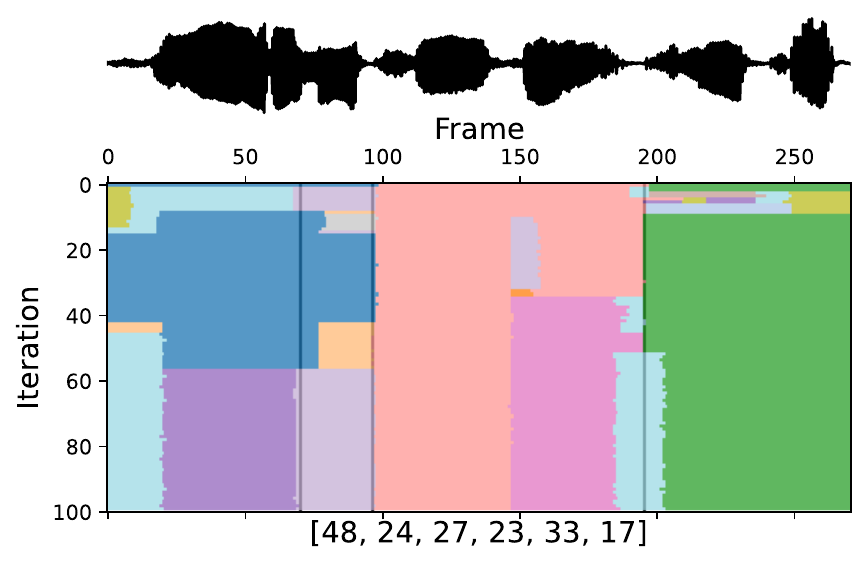} 
    \subcaption{Fixed condition: \\ /\underline{bo:ru} no \underline{chiisai} \underline{omocha}/}
    \footnotesize{(A small toy of a ball.)} 
    \label{fig:ex_2_segment2}
  \end{subfigure}
  \begin{subfigure}[b]{0.33\linewidth}
    \centering
    \includegraphics[width=\linewidth]{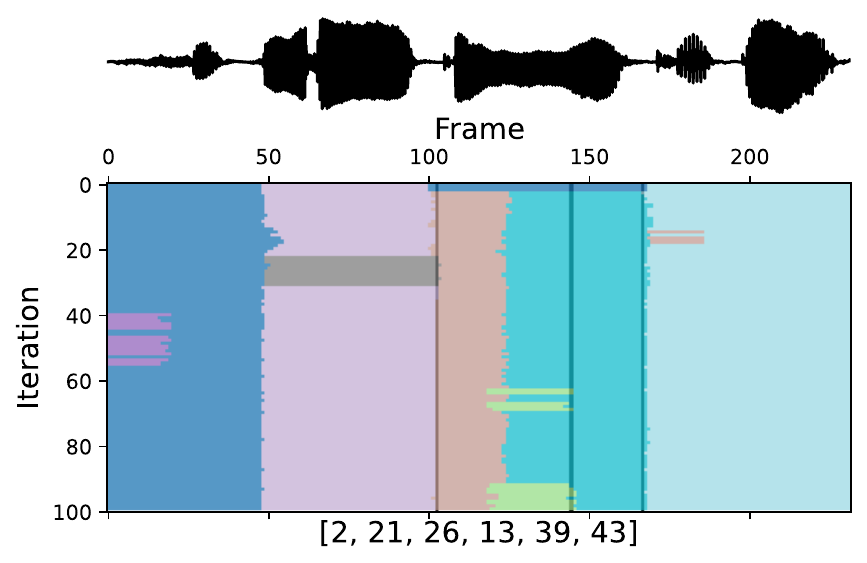} 
    \subcaption{Fixed condition: \\ /\underline{supurei} \underline{kaN} wa \underline{katai}/}\footnotesize{(Spray can is hard.)} 
    \label{fig:ex_2_segment3}
  \end{subfigure}
  \\\\
  \begin{subfigure}[b]{0.33\linewidth}
    \centering
    \includegraphics[width=\linewidth]{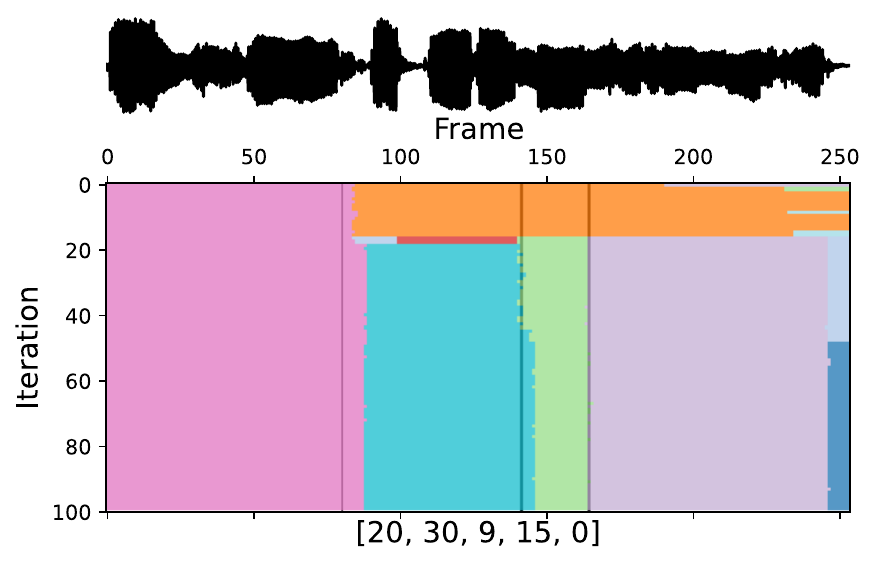} 
    \subcaption{Increase condition: \\ /\underline{oishii} \underline{botoru} no \underline{nomimono}/ }
    \footnotesize{(A bottle of delicious drink.)} 
    \label{fig:ex_2_segment1_i}
  \end{subfigure}
  \begin{subfigure}[b]{0.33\linewidth}
    \centering
    \includegraphics[width=\linewidth]{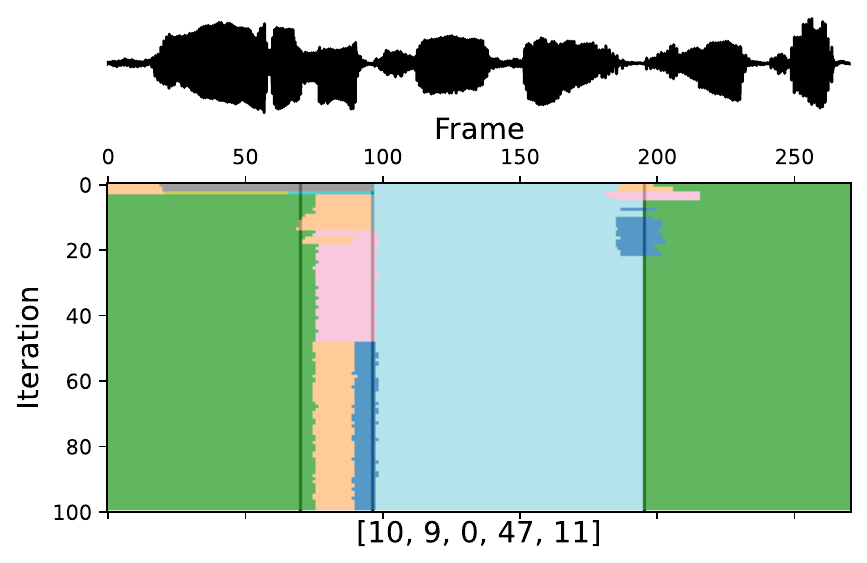} 
    \subcaption{Increase condition: \\ /\underline{bo:ru} no \underline{chiisai} \underline{omocha}/}
    \footnotesize{(A small toy of a ball.)} 
    \label{fig:ex_2_segment2_i}
  \end{subfigure}
  \begin{subfigure}[b]{0.33\linewidth}
    \centering
    \includegraphics[width=\linewidth]{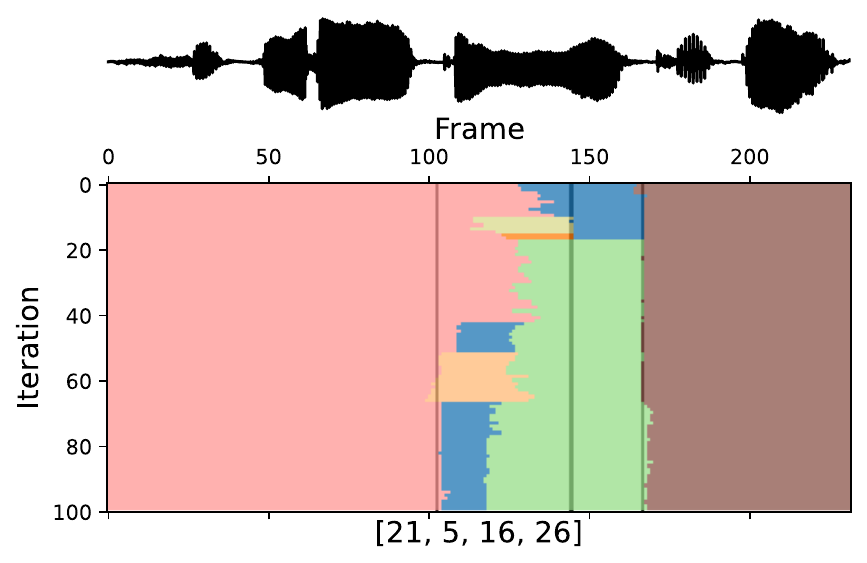} 
    \subcaption{Increase condition: \\ /\underline{supurei} \underline{kaN} wa \underline{katai}/}\footnotesize{(Spray can is hard.)} 
    \label{fig:ex_2_segment3_i}
  \end{subfigure}
  \caption{Examples of word segmentation results (Experiment 2):
  The upper sub-figures (a, b, c) show the results of the co-occurrence DAA (Fixed condition).
  The lower sub-figures (d, e, f) show the results of the co-occurrence DAA (Increase condition).
  }
  \label{fig:segment2}
\end{figure*}

Table~\ref{tab:ex_2_last_ARI} shows the performance of word discovery and categorization at the end of training for each weight value setting.
As in Experiment 1, there was no difference in phoneme performance, so we omitted the evaluation in this experiment.
As a result, the performance was higher when the word modality weights were gradually increased than when they were fixed.

Overall, the co-occurrence DAA{$^\ast2$} of the increase condition had the highest performance.
The increase condition does not use word modality in the early stages of learning, when word discovery is uncertain, but uses it for categorization after some learning has already progressed.
{This suggests that improved categorization performance by the increase condition can be used as co-occurrence cues for lexical acquisition, potentially enhancing word discovery performance.}

Alternatively, the decrease condition, which reduced the word modality weights, slightly decreased the word discovery performance and lowered the categorization performance.
The experimental results support our hypothesis that uncertain word segmentation results in the early learning stages may negatively impact classification.

Figure~\ref{fig:segment2} shows examples of the results of word segmentation of speech.
Figures~\ref{fig:ex_2_segment1}, \ref{fig:ex_2_segment2}, \ref{fig:ex_2_segment3} show examples that even fixed conditions could not be accurately segmented.
In these examples, the increase condition improved word segmentation performance.
In addition, when compared with the results of NPB-DAA in Experiment 1, over-segmentation could be significantly alleviated for words that describe the characteristics of an object.

In the comparison of weighting in SIR, the word modality weights were higher in UR for the increase condition and higher in MI for the fixed condition.
UR worked well with the increase condition because it could focus on more appropriate candidates.
MI worked to retain various candidates, but in some cases, it failed to narrow down the appropriate candidates.
The results suggest that not only MI used in the conventional method, but also UR which is mathematically consistent with the proposed method, can be effective as criteria.

\section{{Discussion and} Conclusion} \label{sec:conclusion}

In this paper, we proposed an unsupervised phoneme and word discovery method that exploits phonological and co-occurrence cues, to imitate the lexical acquisition process of infants using statistical learning.
The main features of the proposed method are the following two points:
(i) It integrates HDP-HLM, a probabilistic generative model for simultaneous phoneme and word discovery, and MLDA, a probabilistic generative model for multimodal object categorization.
(ii) Multimodal sensor observations of image, tactile, and audio stimuli can be simultaneously used as co-occurrence cues for lexical acquisition.
Experimental results showed that the proposed method improved word discovery performance for the entire utterance compared to the existing methods that do not use co-occurrence cues.
These results suggest that the proposed method can find words more accurately than existing methods, insofar as the words express the characteristics of the object.
In addition, increasing the modality weight of the words in the categorization improved the categorization and word discovery performance.

The study focused on the critical elements for language acquisition (i.e., co-occurrence cues) claimed by Safran et al~\cite{saffran2018infant,saffran2020statistical,saffran1996word}. Therefore, perception/cognition and learning, rather than utterance and production, were considered. Future perspectives may involve the utilization of language to utter (reproduce) learned words and sentences on their own. Other essential factors involved in the development of language acquisition include a perceptual reorganization and vocabulary spurt~\cite{Rost2009,Werker2002,Gertner2006,Goldfield1990}. 
Constructing a unified computational model that accounts for multiple developmental stages in language acquisition remains a challenging and unresolved issue.

The proposed method has a limitation in setting the number of object categories. 
However, HDP-MLDA~\cite{nakamura2011multimodal}, which extends MLDA with a nonparametric Bayesian method (specifically, the Dirichlet process), automatically estimates the appropriate number of categories for the data. 
It is easy to extend the MLDA part of the proposed method to HDP-MLDA. 
In the future, the limitation on the number of object categories could be resolved.

Our future study prospects include (i) enabling word discovery that incorporates prosodic cues~\cite{okuda2021prosody} into the proposed method, and (ii) using the acquired words for human-robot interaction and feedback learning through speech synthesis.

In this study, we employed NPB-DAA as an unsupervised phoneme and word discovery method and MLDA as a categorization method using multimodal object information. The essence of the proposed method is the integration procedure that exploits the co-occurrence of phonological and object information in probabilistic generative models.
It does not depend on any particular model as long as it can be represented by a probabilistic generative model.
Therefore, in the future, it will be possible to reconstruct the proposed method based on various speech unit discovery and categorization models.

\section*{Acknowledgment}
This research was partially supported by Grants-in-Aid for Scientific Research on Innovative Areas (16H06569 and 17H06383) and Grant-in-Aid for Scientific Research (A) (21H04904) funded by the Ministry of Education, Culture, Sports, Science, and Technology, Japan.

\bibliographystyle{IEEEtran}
\bibliography{articles,library} %

\clearpage
\appendices

\section{Preliminary Experiments} \label{sec:apdx}

\subsection{Preliminary experiment I: Object categorization performance using accurately written transcript sentences and multimodal sensor information} \label{sec:sub_2}

In this experiment, we confirmed categorization accuracy using multimodal information when the speech dataset was recognized accurately.
First, the modality weight for categorization was determined by categorizing each modality.
Then, the categorization performance when the weight of the word modality was changed was measured.
The result of this experiment can be interpreted as the upper limit (that is, the topline) of the categorization using this dataset.

\subsubsection{Condition}

The word modality uses the word histogram created based on the transcript, which is generated based on the utterance content of the dataset as word sequences.
The multimodal information of an object includes all three modalities of vision, haptic, and audio.
The number of categories and the values of the hyperparameters were the same as those described in Section~\ref{sec:ex_1}.
Here, the weight setting of each modality used a value between $ 0,10,20, \dots, 300 $ for the word modality, $ 100 $ for the vision, $ 100 $ for the haptic, and $ 50 $ for the audio. 
The latter three values were fixed.
The weight value $ 0 $ indicates that categorization was performed while excluding the word modality.
Weight value settings for values other than word modality were set to be proportional to the categorization accuracy, referring to the accuracy when categorization was performed for each modality alone (See Table~\ref{tbl:sub_2_solo_ARI}).



\begin{table}[tb]
    \centering
    \caption{Categorization accuracy when vision, haptic, or audio modalities were used individually.
     (Preliminary experiment I)}    
    \begin{tabular}{cccc}\hline
    \textbf{Modality} & {Vision}  & {Haptic} & {Audio} \\ \hline 
    \textbf{Categorization ACC} & 0.500 & 0.500 & 0.333 \\ \hline
    \end{tabular}
    \label{tbl:sub_2_solo_ARI}
\end{table}

Because the seed value of the random numbers was fixed for the implementation of MLDA, the trial was performed with each weight value.
The Gibbs sampler of MLDA was 1000 iterations.
The number of iterations of the Gibbs sampler was determined in advance by investigating learning iterations in which the categorization converged sufficiently.

\subsubsection{Result}
\begin{figure}[tb]
  \begin{center}
    \includegraphics[width=\linewidth]{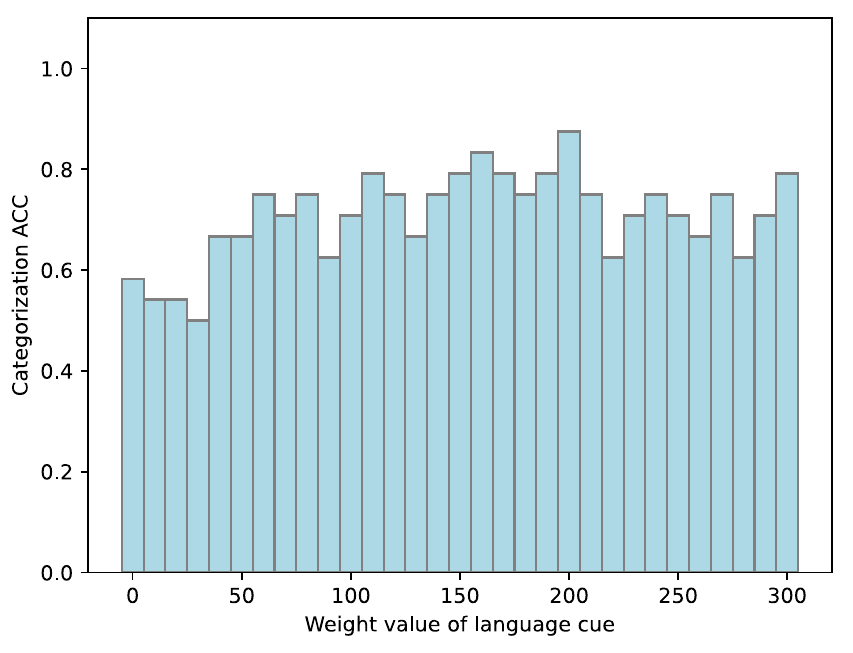}
    \caption{
    Relationship between the weight of word modality and categorization accuracy when using transcription sentences, that is, ground truth. (Preliminary experiment I)
    }
    \label{fig:sub_2_ARI}
  \end{center}
\end{figure}

Figure~\ref{fig:sub_2_ARI} shows the categorization accuracy for each weight of word modality.
Comparing the categorization accuracy for each weight, the accuracy of the weight value $ 0 $ without using the word modality was low, and the linguistic information contributed to the improvement of the categorization accuracy in this dataset.
In addition, the weights of $ 10 $ to $ 30 $ showed almost the same accuracy as when the word modality was not used.
The accuracy tended to increase slightly from $ 40 $ to $ 200 $, and there was no significant change in accuracy after the weight of $ 210 $.
Therefore, it can be inferred that the weight of the word modality was between $ 40 $ and $ 200 $.

\subsection{Preliminary experiment II: Relationship of categorization performance and modality weight using word sequences by NPB-DAA} \label{sec:sub_3} 

In this experiment, we examined the appropriate weight value setting of the word modality using the word sequence estimated by unsupervised learning.
Specifically, the results of the phoneme and word discovery experiments by NPB-DAA were used as MLDA inputs, and the categorization accuracy was compared by setting various weight values.
In addition, by performing categorization using the word sequences estimated in each iteration of NPB-DAA, we investigated how the categorization accuracy changed when using word sequences that had not been sufficiently learned, and how to set the weight value at the time.

\subsubsection{Condition}

The multimodal object and speech dataset is described in Section~\ref{sec:dataset}.
The values of the hyperparameters and the modality weights for object categorization were the same as those described in Section~\ref{sec:sub_2}.
The iteration numbers for the Gibbs sampler of MLDA and NPB-DAA were 1000 and 100, respectively.
Ten trials were performed by NPB-DAA for each modality weight setting.

\subsubsection{Result}

\begin{figure}[tb]
  \begin{center}
    \includegraphics[width=\linewidth]{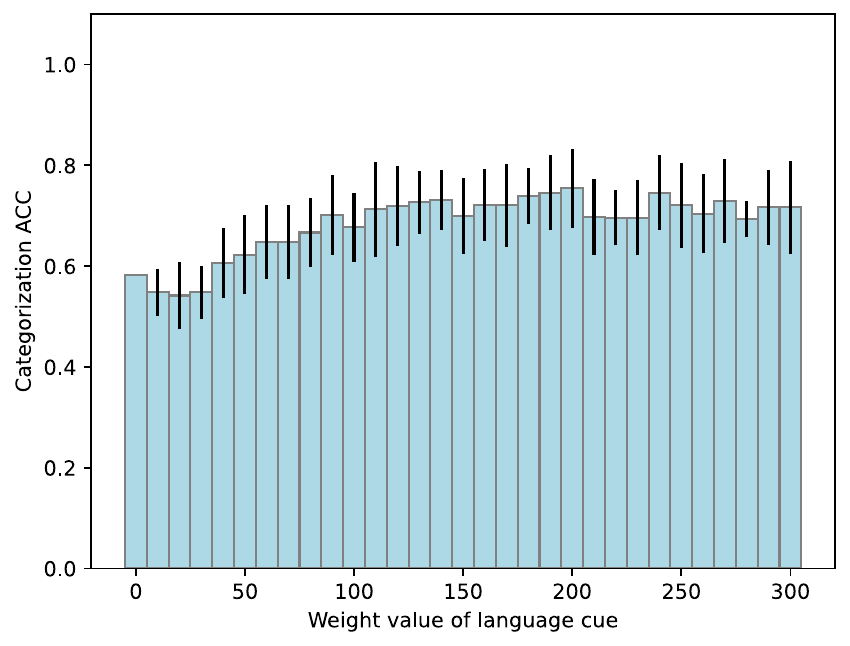}
    \caption{
    Relationship between the weight of word modality and categorization accuracy when using word sequences estimated by NPB-DAA. (Preliminary experiment II)
    }
    \label{fig:sub_3_last_ARI}
  \end{center}
\end{figure}

\begin{figure}[tb]
  \begin{center}
    \includegraphics[width=\linewidth]{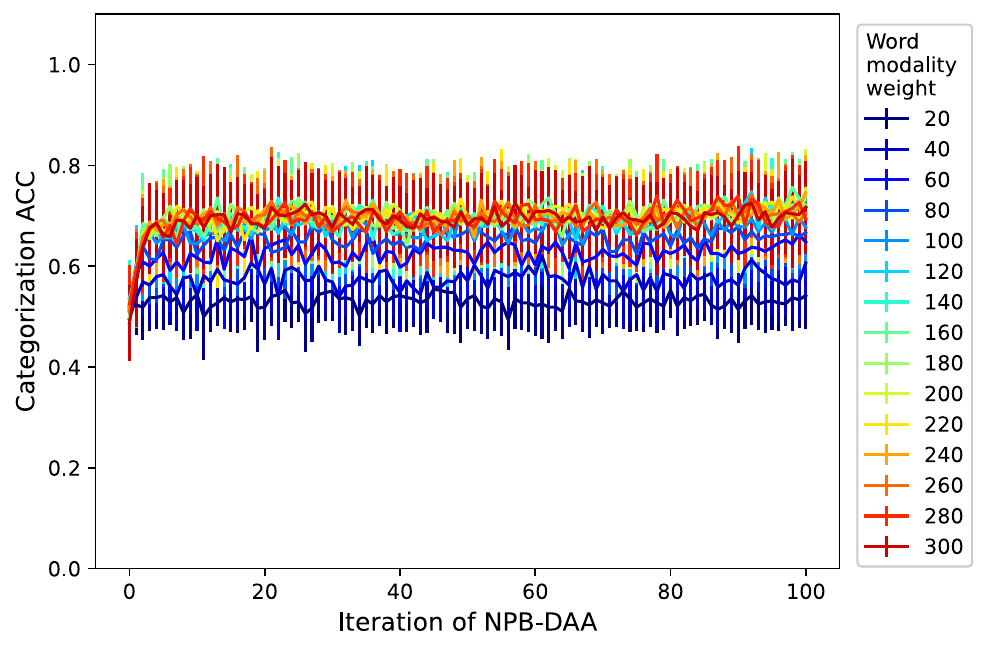}
    \caption{
    Relationship of categorization accuracy between the number of learning iterations of NPB-DAA and the weight of word modality. (Preliminary experiment II)
    }
    \label{fig:sub_3_trans_ARI}
  \end{center}
\end{figure}

Figure~\ref{fig:sub_3_last_ARI} shows the accuracy of object categorization for each word modality weight.
Figure~\ref{fig:sub_3_trans_ARI} shows the changes in the accuracy of object categorization for each word modality weight when using word sequences estimated by NPB-DAA for each iteration of NBP-DAA.
From Figures~\ref{fig:sub_2_ARI} and \ref{fig:sub_3_last_ARI}, 
it can be inferred that the weight of word modality of 40 to 200 is appropriate because the relationship between the weight of the word modality and the categorization accuracy was similar to that of Section~\ref{sec:sub_2}.
Furthermore, from Figure~\ref{fig:sub_3_trans_ARI}, 
the number of iterations of NPB-DAA did not significantly affect the categorization accuracy for each weight of word modality, except for the initial stage of NPB-DAA.

\end{document}